\documentclass[lettersize,journal, ]{IEEEtran}
\usepackage{amsmath,amsfonts}
\usepackage{algorithmic}
\usepackage{array}
\usepackage{textcomp}
\usepackage{url}
\usepackage{verbatim}
\usepackage{graphicx}
\def\BibTeX{{\rm B\kern-.05em{\sc i\kern-.025em b}\kern-.08em
    T\kern-.1667em\lower.7ex\hbox{E}\kern-.125emX}}
\usepackage{balance}
\usepackage{verbatim}
\usepackage{amsfonts}
\usepackage{amssymb}
\usepackage{array}
\usepackage{textcomp}
\usepackage{url}
\usepackage{verbatim}
\usepackage{graphicx}
\def\BibTeX{{\rm B\kern-.05em{\sc i\kern-.025em b}\kern-.08em
    T\kern-.1667em\lower.7ex\hbox{E}\kern-.125emX}}
\usepackage{balance}
\usepackage{verbatim}
\usepackage{amsfonts}
\usepackage{amssymb}
\usepackage{cite}
\usepackage{soul}
\usepackage{setspace}
\usepackage{graphicx}
\usepackage{psfrag}
\usepackage{array}
\usepackage{epstopdf}
\usepackage{authblk}
\usepackage{graphicx} 
\usepackage{amsthm} 
\usepackage{lipsum}
\usepackage{verbatim} 
\usepackage{authblk}
\usepackage{mathtools}
\usepackage{cuted}
\usepackage[lined,boxed,ruled]{algorithm2e}
\usepackage{booktabs}
\usepackage{subfigure}
\usepackage[font=small]{caption} 
\usepackage{ifthen}
\usepackage[usenames]{color}

\newtheorem{Definition}{Definition}

\newtheorem{Lemma}{Lemma}

\newtheorem{Proposition}{Proposition}
\newtheorem{Remark}{Remark}

\DeclareMathOperator*{\argmax}{argmax}

\newboolean{showaircomp}
\setboolean{showaircomp}{False} 
\usepackage{bm}
\usepackage{verbatim}
\usepackage{amsfonts}
\usepackage{amssymb}
\usepackage{stfloats}
\usepackage{cite}
\usepackage{graphicx}
\usepackage{psfrag}
\usepackage{subfigure}
\usepackage{amsmath}
\usepackage{array}
\usepackage{epstopdf}
\usepackage{authblk}
\usepackage{graphicx} 
\usepackage{amsthm}         
\usepackage{lipsum}
\usepackage{verbatim} 
\usepackage{authblk}
\usepackage{mathtools}
\usepackage{cuted}
\usepackage{authblk}
\usepackage{booktabs}
\usepackage{subfigure}
\usepackage{siunitx}
\usepackage{mathtools}
\usepackage{soul}

\begin{document}

\title{Communication Efficient Cooperative Edge AI via Event-Triggered Computation Offloading}

\author{{You~Zhou,~\IEEEmembership{Graduate Student Member,~IEEE}, 
{Changsheng You,~\IEEEmembership{Member,~IEEE}}, 
and {Kaibin Huang,~\IEEEmembership{Fellow,~IEEE}}}

\thanks{Y. Zhou and K. Huang are with the Department of Electrical and Electronic Engineering (EEE), The University of Hong Kong (HKU), Hong Kong SAR, China (Email: zhouyou@eee.hku.hk, huangkb@eee.hku.hk). C. You is with the Department of EEE, Southern University of Science and Technology (SUSTech), China (Email: youcs@sustech.edu.cn). The corresponding author is K. Huang.}}

\maketitle

\begin{abstract}
Rare events, despite their infrequency, often carry critical information and require immediate attentions in mission-critical applications such as autonomous driving, healthcare, and industrial automation. The data-intensive nature of these tasks and their need for prompt responses, combined with designing edge AI (or edge inference), pose significant challenges in systems and techniques. 
Existing edge inference approaches often suffer from communication bottlenecks due to high-dimensional data transmission and fail to provide timely responses to rare events, limiting their effectiveness for mission-critical applications in the \emph{sixth-generation} (6G) mobile networks.
To overcome these challenges, we propose a channel-adaptive, event-triggered edge-inference framework that prioritizes efficient rare-event processing. Central to this framework is a dual-threshold, multi-exit architecture, which enables early local inference for rare events detected locally while offloading more complex rare events to edge servers for detailed classification. To further enhance the system’s performance, we developed a channel-adaptive offloading policy paired with an online algorithm to dynamically determine the optimal confidence thresholds for controlling offloading decisions. The associated optimization problem is solved by reformulating the original non-convex function into an equivalent strongly convex one.          
Using deep neural network classifiers and real medical datasets, our experiments demonstrate that the proposed framework not only achieves superior rare-event classification accuracy, but also effectively reduces communication overhead, as opposed to existing edge-inference approaches.
\end{abstract}

\section{Introduction}\label{Sec:Introduction}

 The \emph{sixth-generation} (6G) mobile networks are envisioned to enable a wide range of intelligent applications by integrating edge \emph{artificial intelligence} (AI) and network sensing. By combining the computational power of edge AI with the pervasive sensing capabilities of networked devices, these networks aim to provide intelligent real-time services across diverse scenarios \cite{liu2022integrated, zeng2024knowledge,wang2024spectrum}. Among the critical applications of 6G networks, the timely detection and processing of rare but high-impact events, such as security breaches or medical emergencies, stands out as a key challenge. Rare events, though infrequent, often carry critical information that demands immediate attention and efficient resource allocation. However, the data-intensive nature of AI-driven applications, coupled with the need for energy-efficient processing, complicates the ability to respond promptly to these events within constrained resources. To address these challenges, our work introduces the framework of \emph{dynamic cooperative inference}. This framework facilitates adaptive collaboration between devices and edge servers to split the execution of AI algorithms, specifically targeting the efficient detection and processing of rare events. By leveraging an event-triggered approach, the framework significantly minimizes unnecessary radio-access transmissions, alleviating communication bottlenecks, and enabling the resource-efficient deployment of AI-powered sensing systems in 6G networks.

Rare events, offer unique opportunities to enhance the split inference framework by prioritizing high-impact data and optimizing resource allocation. Split inference, a popular architecture for edge AI, partitions an inference task between a device and a server to overcome the device's resource constraint. This design offloads complex computations to edge servers while allowing devices to handle initial processing, significantly reducing the latter's computational burdens and also the communication overhead. Research has shown that split inference improves the inference accuracy under dynamic channel conditions \cite{li2019edge}, reduces communication costs \cite{yan2022optimal,wang2024ultra}, and supports hardware-limited devices \cite{lee2023wireless}. Additionally, split inference takes advantage of the sparsity of intermediate data by incorporating feature compression to reduce communication overhead \cite{shao2020bottlenet++}. However, this method remains insufficient device's transmission of high-dimensional intermediate features still create a substantial communication bottleneck. This poses a critical issue in rare-event scenarios where high-priority data require immediate and precise transmission. Joint source-and-channel coding frameworks improve robustness in noisy environments \cite{jankowski2020joint}, but their fixed transmission strategies does not recognize the urgency of rare events, potentially delaying critical responses. Similarly, advanced solutions such as integrated sensing and communication (ISAC) frameworks \cite{wen2023edgeai} and task-oriented over-the-air computation (AirComp) \cite{wen2023task} address communication constraints in multi-device systems but lack the capability to dynamically prioritize rare, high-impact data. Similarly, existing methods as batching and early exiting \cite{liu2023resource} optimizes throughput but fails to prioritize critical information, treating all data equally. These limitations restrict the system's ability to deliver timely and efficient responses in critical scenarios. Addressing this gap requires integrating event-triggered mechanisms to detect and prioritize rare events, allowing systems to reduce unnecessary transmissions while ensuring rapid responses. By incorporating such mechanisms, split inference can overcome communication bottlenecks and fully realize its potential in handling high-stakes, rare-event scenarios.

Computation offloading, with split inference as a specialized branch for task classification, offloads computationally intensive tasks from resource constrained devices to resourceful servers, thereby overcoming the former's limitation in storage-and-computation capacities and energy\cite{mao2017survey,taomeixia2019optimal,you2018asynchronous,geofferyli2021joint,wang2016mobile,mao2016dynamic,hu2018joint,niyato2021fast}. Recent research has developed energy-efficient offloading controllers for asynchronous systems with dynamic event arrivals \cite{you2018asynchronous}, resource allocation techniques for satellite-aerial edge networks \cite{geofferyli2021joint}, and dynamic voltage and frequency scaling (DVFS) methods to adjust computational speed based on demand \cite{wang2016mobile}. However, despite these advances, existing approaches overlook the unique demands and potential efficiency gains associated with rare, high-impact events. Specifically, traditional models typically distribute the workload of all events uniformly between devices and servers without favouring rare events in resource allocation and as a result, fail to meet their mission critical requirements\cite{cao2018joint}. Similarly, while techniques such as Lyapunov optimization adapt offloading decisions to dynamic network states to minimize execution delay, they fail to differentiate between routine and critical tasks, potentially delaying responses to high-priority events\cite{mao2016dynamic}. In addition to their inability to prioritize rare events, many existing methods struggle with excessive communication overhead. Offloading techniques such as UAV trajectory optimization and user scheduling aim to reduce latency, but they do not address the high-dimensional data generated during rare events, which often overwhelms network capacity\cite{hu2018joint}. Similarly, lagrange coded computing frameworks enable fast and secure offloading, but their fixed coding and transmission strategies are inefficient for handling the sporadic nature of rare events\cite{niyato2021fast}. Similar limitation exist for techniques such as resource optimization in satellite-aerial edge networks focusing preplanned allocations\cite{geofferyli2021joint}. In summary, event-triggered offloading presents an opportunity to overcome the communication bottleneck of existing techniques. By selectively focusing on critical, high-impact events, computational workload and network congestion can be significantly reduced, leading to more efficient resource allocation and meeting the requirements of mission critical applications. 

Building on the preceding discussion, it is evident that rare events and their associated sparse access patterns are important considerations for optimizing resource utilization in 6G communication systems. Sparse access, characterized by sporadic device activity, has been extensively studied in the context of Internet of Things and machine-type communications networks for its potential to enhance the communication efficiency\cite{liu2018massive}. For instance, compressed sensing techniques enable signal reconstruction from fewer measurements to reduce computational overhead at sensor nodes \cite{qin2018sparse}. On the other hand, grant-free access schemes exploit sparse traffic patterns to minimize contention delays in random access protocols \cite{liu2018sparse}. Additional advancements, such as integrated sensing and communication frameworks \cite{niuzhisheng2021sensing} and adaptive anomaly detection methods \cite{tonyquek2021adaptive}, further enhance the multi-access efficiency by leveraging the sporadic nature of device activity. Despite these advancements, existing sparse access techniques still have significant limitations. In particular, many existing methods for rare event detection are confined to primitive tasks, such as monitoring simple environmental parameters (see e.g., \cite{dereszynski2011spatiotemporal}, which do not address the complex sensing requirements of 6G networks \cite{zhu2023pushing}. Furthermore, while sparse access research has made substantial progress, its integration with channel dynamics remains insufficiently explored. This is useful given the unpredictable nature of rare events and the variability of network and channel conditions. For instance, systems like collision avoidance in autonomous driving critically depend on the seamless integration of rare-event detection with channel-adaptive mechanisms to ensure timely and accurate responses under dynamic network environments \cite{bojarski2016end}. These gaps underscore the necessity of a channel-adaptive, event-triggered system capable of dynamically adjusting computation and communication based on real-time channel conditions and event criticality.

To address the issue, we consider an event-triggered system designed for rare-event detection in edge AI applications. The objective of this system is to maximize end-to-end (E2E) rare event classification accuracy while operating within constrained communication and computation resources. Achieving this objective requires a framework that integrates efficient event detection, adaptive inference, and dynamic offloading techniques. In this work, we present an event-triggered framework that combines \textit{early exiting} (see e.g., \cite{exiting}) and channel-adaptive computation offloading. This framework dynamically adjusts computation and communication based on the real-time channel state, enabling efficient communication inference. The key contributions and findings are summarized as follows.

\begin{itemize}
     \item \textbf{Design of Dynamic Inference Architecture}: We propose a novel event-triggered cooperative inference framework, refered to as co-inference, featuring a dual-threshold, early-exiting architecture. This system enables efficient local binary tail-event detection while offloading rare, high-impact events to the server for refined multi-class classification. The dual thresholds create an uncertainty region where events that cannot be confidently classified continue to subsequent blocks for a decision with better confidence. This flexibility allows highly confident events to exit early for rapid local inference, while uncertain events are processed further, and complex rare cases are offloaded to the server for detailed analysis. By balancing the event's missing target probability and offloading probability, the framework effectively balance the tradeoff between communication and computation while preserving high inference accuracy, making it particularly suitable for long-tail distributed events.
     \item \textbf{Channel-Adaptive Offloading Policy Design with Threshold Optimization}: To enhance E2E classification performance under energy and communication constraints, we develop a channel-adaptive offloading policy and an algorithm to dynamically determine the optimal dual-thresholds on confidence for rare event offloading. This policy enables real-time rare event offloading, effectively balancing local computation with offloading workload to adapt to varying channel conditions. By transforming the non-convex optimization problem into a strongly convex equivalence, our approach maximizes rare event classification accuracy under constraints on energy consumption and communication overhead.
     \item \textbf{Experiments}: The effectiveness of our proposed event-triggered co-inference system is demonstrated using real medical datasets across various CNN architectures (i.e., ShuffleNet and MobileNet \cite{ma2018shufflenet, sandler2018mobilenetv2}). The results reveal a significant reduction on both computation and communication overhead while consistently outperforming traditional split-inference techniques in term of classification accuracy.
   \end{itemize}

The remainder of this paper is organized as follows. The dynamic co-inference system models and metrics are introduced in Section II. An overview of the co-inference architecture is demonstrated in Section III. The dual-threshold multi-exit model and tradeoffs during event offloading are presented in Section IV, while the offloading policy is introduced and the dual-threshold for rare event classification is optimized in Section V. Section VI reports the experimental results, followed by concluding remarks in Section VII.
\begin{figure}[tpb]
      \centering
      \hspace{-0.5cm} 
	  \includegraphics[scale=0.15]{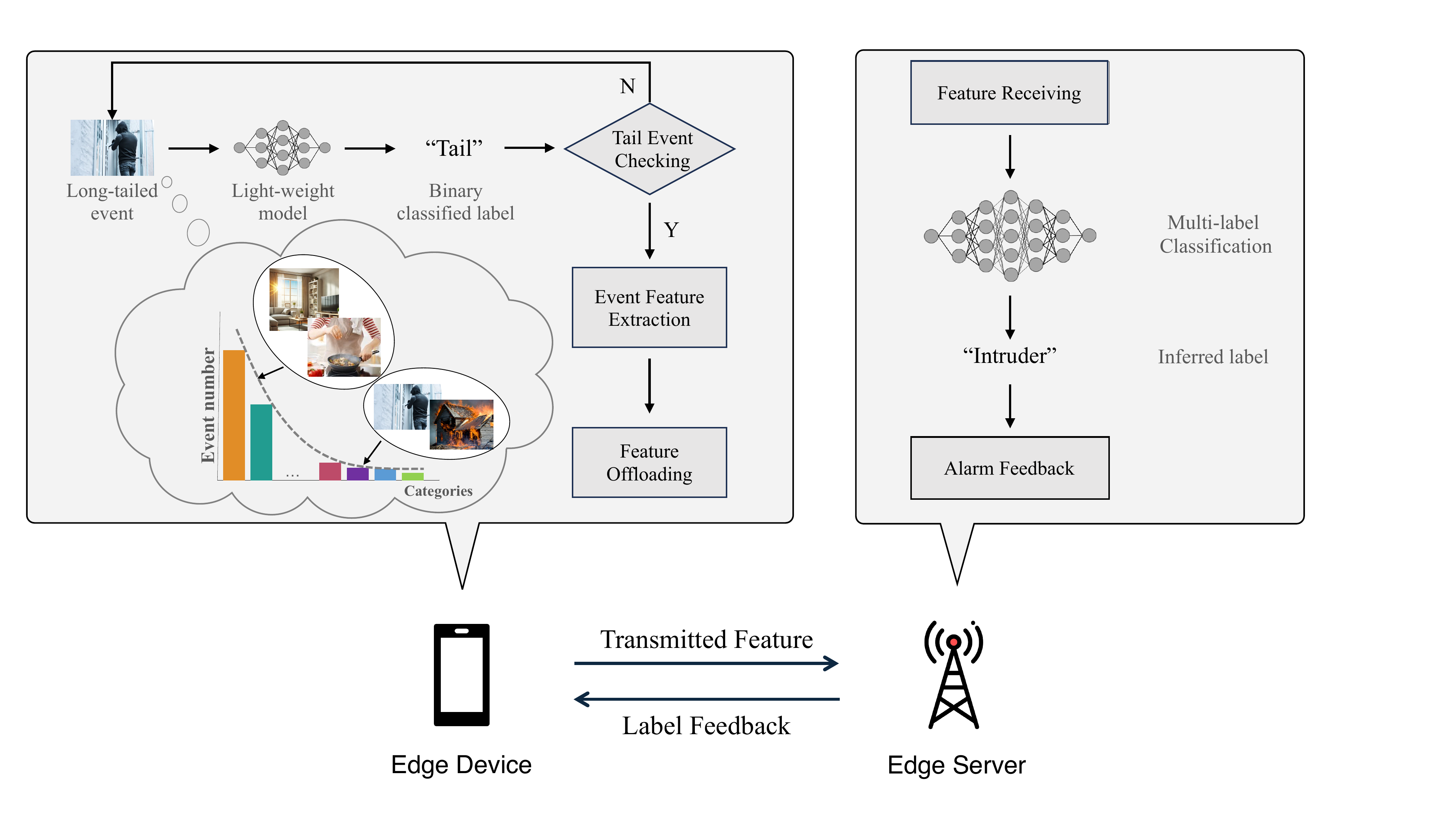}
      \caption{Edge co-inference system.}
      \vspace{-3mm}
\end{figure}
\section{Models and Metrics}\label{Sec:ModelandMetrics}
We consider an edge co-inference system, depicted in Fig. 1, where an edge server collaborates with a mobile device to perform inferences for a sequence of events in the local device's event queue. The event queue follows a first-in-first-out (FIFO) order. Unlike existing works that primarily focus on balanced data distribution, this paper addresses the practical scenario of unbalanced data. Specifically, we focus on a \textit{long-tailed distribution} model, characterized by a majority of head (normal) events and a minority of tail (rare) events\cite{longtail}. The specific models and metrics are described in the sub-sections.

\subsection{Inference Model} 
\subsubsection{CNN Classification with Early Exiting}
We adopt the CNN model, commonly used for image recognition and processing, which includes multiple convolutional (\textit{CONV}) layers, pooling layers, rectified linear unit (ReLU) correction layers, and fully-connected (\textit{FC}) layers. Let $\mathbf{X}\in\mathbb{R}^{N\times L_\text{h}\times L_\text{w}}$ denote the tensor of intermediate feature maps received at the edge server from an input event. Here, $N$ represents the number of feature maps, each with height $L_\text{h}$ and width $L_\text{w}$. Given the ground-truth label for an event $m$, denoted as $\hat{\ell}_m$, the server can compute the posterior $\text{Pr}(\ell_m|\mathbf{X})$ using the forward-propagation algorithm and estimate the label $\hat{\ell}_m$ by maximizing the posterior probability, i.e., $\hat{\ell}_m = \argmax{\ell_m} \text{Pr}(\ell_m|\mathbf{X})$.
To address the high communication cost caused by large original input data (e.g., 3D images and films), we employ an early-exiting model, allowing for classification at earlier stages instead of traversing the entire CNN structure. Following the standard architecture in the literature, the early-exiting model at the edge server divides the CNN model into $N$ consecutive blocks, each followed by a classifier that can predict the label using the extracted features, as illustrated in Fig. 2. An intermediate classifier is a simplified neural network with fewer layers compared to the conventional CNN block.
\subsubsection{Local Inference Energy Consumption}
In the context of CNNs, the local inference energy is influenced by both memory access and arithmetic operations. However, memory access typically consumes significantly more power compared to arithmetic operations\cite{ma2018shufflenet}. Therefore, we focus on characterizing energy consumption and latency based on memory access operations. Given one CNN contains 
$N$ blocks, the local energy consumption can be calculated as follows \cite{mao2016dynamic}:
\begin{equation}\label{localenergy}
E_\text{loc}(N) = \sum_{i=1}^{N}S^{\text{mem}}_{i}\varrho,
\end{equation}
where $S^{\text{mem}}_{i}$ denotes the number of memory access operations at the $i$-th block, and $\varrho$ is the energy per memory access operation.

\begin{figure}[tpb]
      \hspace{+0.5cm}
	  \includegraphics[scale=0.25]{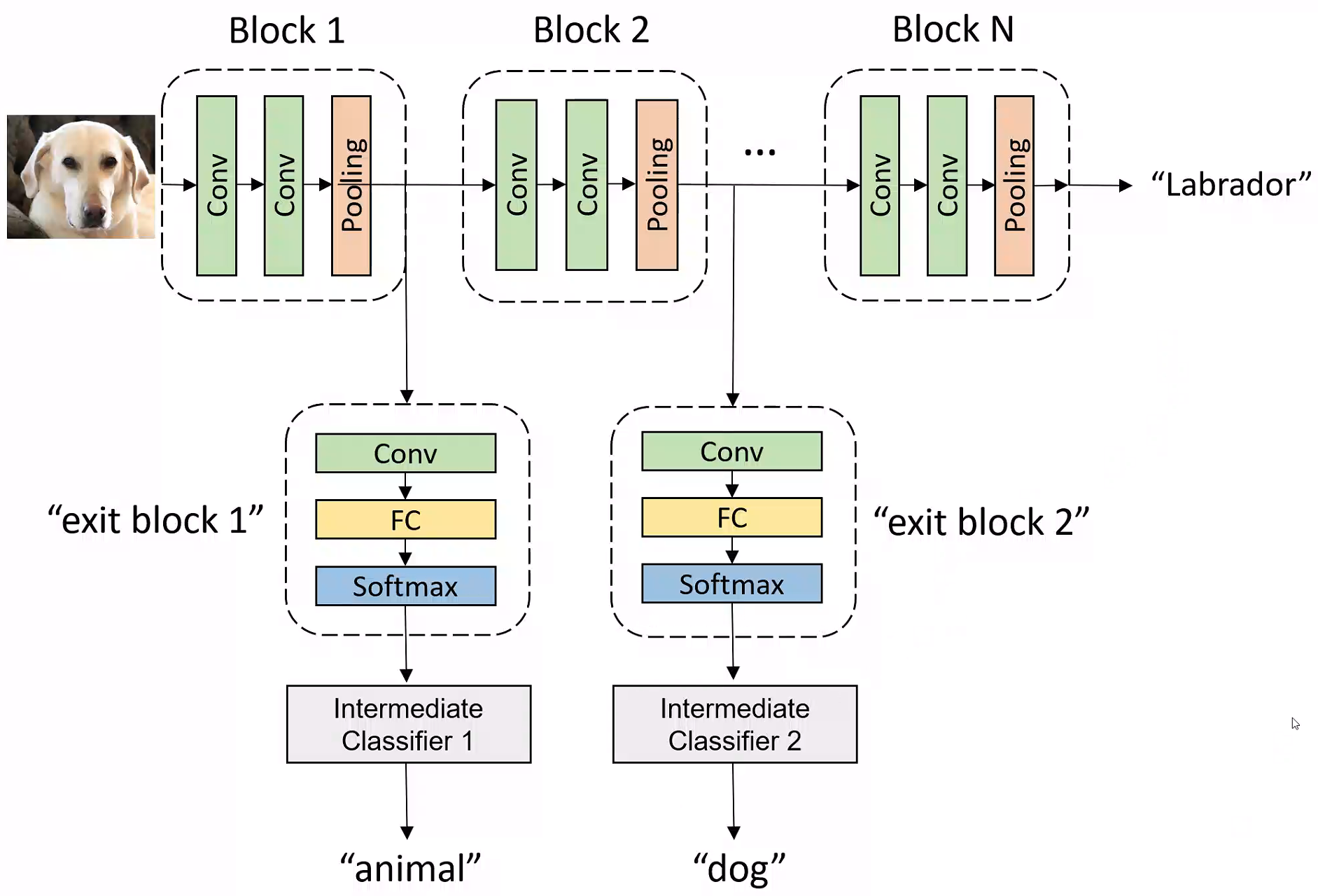}
      \caption{Illustration of a backbone model with early exiting.}
      \vspace{-3mm}
\end{figure}
\vspace{-3mm}
\subsection{Communication Model}
In our extended modeling of energy consumption for co-inference scenarios, we account not only for the computational energy required by the local device but also for the energy expended during the transmission of uplink features. The energy required to offload is primarily influenced by the size of the extracted feature maps being transmitted from the local device to the edge server. Representing the size of the feature input data as $D$ and the transmit power of the mobile device as $P_{\text{tr}}$, the energy consumed during the feature offloading can be quantified as:
\begin{equation}\label{offloading energy}
\begin{split}
      E_{\text{off}}=P_{\text{tr}}\frac{D}{R_\text{tr}},
\end{split}
\end{equation}
where $R_{\text{tr}}$ represents the transmission rate for offloading events. The transmission rate depends on the available bandwidth $B$, channel gains $G$, and noise power $\sigma^2$, and can be computed using the equation\cite{taomeixia2019optimal}:
\begin{equation}
    R_\text{tr} = B\log(1+ \text{SNR}).
\end{equation}
In practical scenarios, $P_{\text{tr}}$ is often maintained constant to simplify power management in wireless devices. During each coherent time, the channel conditions can vary significantly due to factors such as environmental changes, mobility of the transmitter and receiver, and other dynamic elements affecting the channel. Consequently, the SNR during each coherent time may differ, leading to variations in the transmission rate 
$R_\text{tr}$ and offloading energy $E_\text{off}$.

\subsection{Performance Metrics}
\subsubsection{Classification Accuracy}
The classification accuracy is defined as the proportion of correctly categorized events among the whole testing dataset. Its empirical accuracy typically approximates the likelihood that an uploaded event is accurately identified. Let $\widehat{x}_m, x_m$ denote the predicted label and the ground truth label of event $m$, respectively. The event $m$ is accurately inferred if $\hat{x}_m = x_m$.

\subsubsection{Mobile Energy Consumption}
The energy consumption of the mobile device is quantified from the initiation of local inference until the completion of offloading the intermediate features to the server. In this context, the overall mobile energy consumption consists of two distinct components: 1) the energy consumed by local computation during partial inference, denoted as $E_\text{loc}$, and 2) the energy consumed during the offloading process, denoted as $E_\text{off}$.

\section{Overview of Dynamic Co-inference}\label{Sec:OverviewofDynamicCo-inference}

To accurately detect tail events and minimize the device's overall energy consumption, we propose a two-stage dynamic co-inference model, as depicted in $\text{Fig. 1}$. A lightweight CNN is deployed on the local device for tail event detection, while a deep CNN on the server handles multi-class classification of the detected tail events. The system operates through cooperating two inference models and the controller for dynamic co-inference adaptes the model by making offloading decisions. The advantages of the proposed two-stage dynamic co-inference model are two-fold. First, it enables a new event-triggered scheme to significantly reduce the communication overhead for server inference since a majority of head events can be successfully detected at the mobile device by properly designing the early exiting model. Second, the early existing model also greatly reduces the computation loads at the mobile device since the head events do not need to traverse all CNN block layers for detection. 
In this instance, the precise configuration of the intermediate classifier, particularly the threshold settings, is crucial for accurately detecting the event and significantly influences the model's performance. The detailed operations are demonstrated as follows.

\subsection{Model 1 - Local Inference}
The first stage aims to distinguish head and tail events by using the intermediate classifiers of the local device model. Specifically, the device will first make a binary classification to label the input event $m$ into either the head or tail class. By applying the \textit{early exiting} technique, the intermediate classifier $C_n(\beta_{\text{u}}, \beta_{\ell})$ is located on the device model's $n$-th block for detection. 
In particular, the intermediate classifier at each block $n$ will predict the input event label by comparing the event confidence score at block $n$ with two confidence thresholds, namely, the upper threshold $\beta_{\text{u}}$ and the lower threshold $\beta_{\ell}$. The two thresholds are set commonly for all intermediate classifiers, which will be detailed in Section~{IV}. After going through the block $n$, if the event $m$'s output confidence $C_{n(m)}$ is between two thresholds, i.e., $\beta_{\ell}\leq C_{n(m)}\leq \beta_{\text{u}}$, the intermediate feature will be passed to the next block for further classification. In stage 1, our target is to determine the optimal threshold set of intermediate classifiers for minimizing the tail event's missing probability.

\subsection{Model 2 - Event-Triggered Computation Offloading}
If the event $m$'s confidence score at block $n$ is in the detectable region, i.e., $C_{n(m)}>\beta_{\text{u}}$ or $C_{n(m)}<\beta_{\ell} $, the event can be classified as follows.: 
\begin{enumerate}
  \item If $C_{n(m)}<\beta_{\ell}$, then the event is labeled as a head event, i.e., $\widehat{x}_m\rightarrow \mathcal{L}_\text{head}$. In this case, the inference result will be output immediately after this block, and the co-inference model will be restarted to detect a new event.
  \item Otherwise, if $C_{n(m)}>\beta_{\text{u}}$, the event is detected as a tail event, i.e., $\widehat{x}_m\rightarrow \mathcal{L}_\text{tail}$. In this case, the classifiers in the subsequent local blocks will be set inactive. However, the lightweight local CNN model may not perform well in multi-class classification and could also lead to significant energy consumption. To address this issue, we propose to offload the detected tail event's features to the server. To minimize the device's total energy consumption, an online algorithm for event offloading decision will be designed in Section~{V}.
  
\end{enumerate}

\begin{figure}[tpb]\label{dualthresholdsfigure}
      
      \hspace{-0.5cm} 
      \includegraphics[scale=0.15]{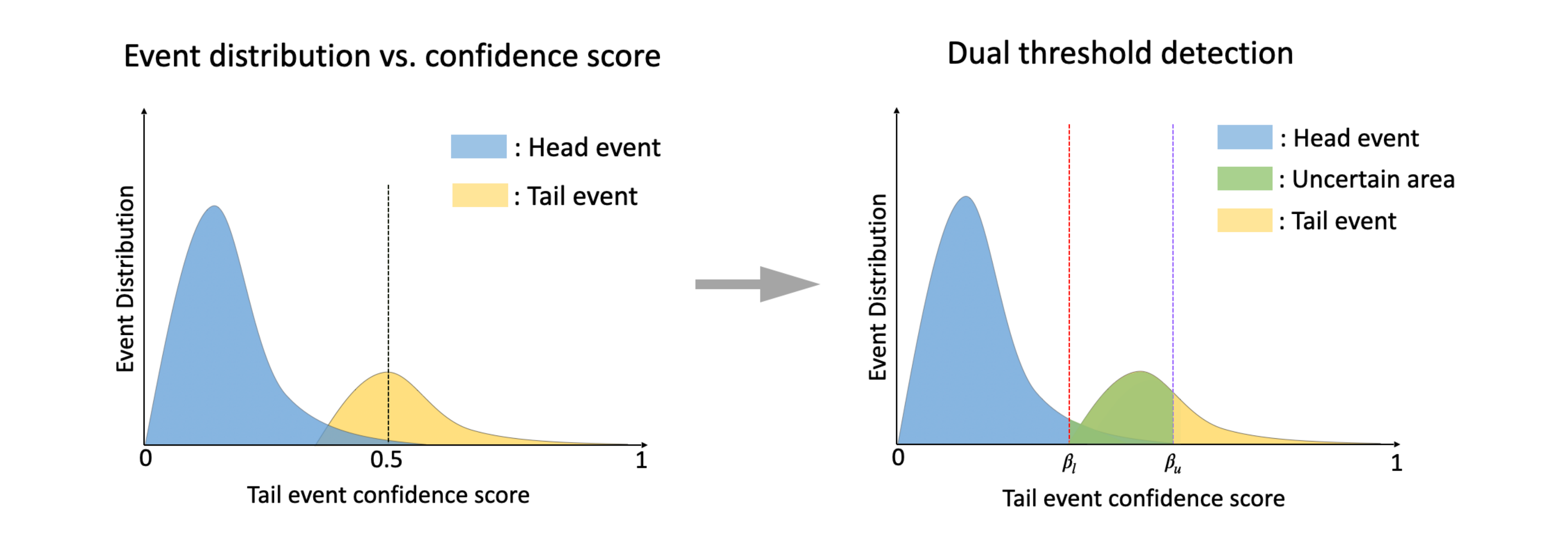}
      \caption{Dual-threshold confidence aware inference progress.}
      \vspace{-3mm}
\end{figure}
\section{Dual-threshold Based Event Detection - Scheme and Tradeoff}\label{DynamicCo-inference:event-Triggered Offloading}
 In this section, we present a new confidence-aware detector based on dual threshold evaluation. We then show the existence of a fundamental tradeoff for the scheme between the (tail-event) missing probability and offloading probability.
 
 \vspace{-2mm}
\subsection{Event Detection Scheme}
Generally, an event classifier model generates a confidence score for each event to indicate its confidence level as a tail event. Given an imbalanced data distribution, the event detection by applying a single threshold on confidence can cause the issue of high (tail-event) missing probability or (head-event) false-alarm probability. As illustrated in the left part of Fig. 3, the distributions of head and tail events are plotted against the tail event confidence score. When a single threshold is set at the traditional value of 0.5, events with a confidence score above 0.5 are classified as tail events, and those below 0.5 are classified as head events. However, this approach results in significant misclassification: a portion of head events is incorrectly identified as tail events, and a substantial portion of tail events is misclassified as head events. In statistics, this issue can be addressed using dual thresholds, with one detecting head events and the other for tail events as proposed in \cite{dual-threshold}. Inspired by this method, we propose a new event detector targeting the backbone model with early exiting (see the right part of Fig. 3), where dual thresholds are applied to confidence scores generated by intermediate classifiers. Before delving into the detailed design, it is necessary to introduce an essential metric for tail-event detection, known as the \emph{confidence score}, which is defined below.
\begin{Definition}\rm
    Given the $m$-th input event enters the $n$-th exit block of the event-detection model. Let two neuron outputs of the exit block $m$ be denoted as $f^{\text{tail}}_{n(m)}$ and $f^{\text{head}}_{n(m)}$, which are associate with the head and tail events, respectively. The confidence score output by the SoftMax function for detecting a tail event is defined as\cite{zhang2021learning}:
    \begin{equation} \label{confidence score}
    \begin{split}
   C_{n(m)}^\text{tail} = \frac{e^{f^{\text{tail}}_{n(m)}}}{e^{f^{\text{tail}}_{n(m)}}+e^{f^{\text{head}}_{n(m)}}}.
    \end{split}
    \end{equation}
\end{Definition}
Next, we apply dual thresholds on the confidence scores of exit blocks sequentially (in the order of increasing depth) until the event is classified confidently. To this end, let the thresholds be denoted as $\beta_{\ell}$ and $\beta_{\text{u}}$ with $0<\beta_{\ell}<\beta_{\text{u}} < 1$. The procedures for detecting head/tail events are described in the sequel. 
 
\subsubsection{Head Event Detection} Consider the $n$-th exit block of the event-detection sub-model and its confidence score defined in \eqref{confidence score}. The condition  $C_{n(m)}^\text{tail}< \beta_{\ell}$ indicates a head event since the opposite has a low probability. Furthermore, the sequential detection requires that $ \beta_{\ell}\leq C_{n(m)}^\text{tail} \leq\beta_{\text{u}}$, $1\leq n\leq N-1$. The validation of the above conditions can be articulated through the evaluation of the following asymptotic indicator function:
\begin{equation}\label{head}
\small
    \mathbb{I}^\text{head}_{n}(m,\beta_{\ell},\beta_\text{u}) = \mathbb{\sigma}(\beta_{\ell}-C_{n(m)}^\text{tail})\prod \limits_{k=1}^{n-1}\mathbb{\sigma}(\beta_{\text{u}}-C_{k(m)}^\text{tail})\mathbb{\sigma}(C_{k(m)}^\text{tail}-\beta_{\ell}),
\end{equation}
where $\sigma(x)$ denotes the Verhulst logistic function given as:
\begin{equation}
\begin{split}
    \sigma(y) = \frac{1}{1+e^{- \alpha y}},  \quad   \alpha\rightarrow\infty.
\end{split}
\end{equation}
Note that as $ \alpha\rightarrow\infty$, the logistic function converges to the Heaviside step function, namely that  $\sigma(y)$ is 1 if $y$ is non-negative or else 0. Therefore, $\mathbb{I}^\text{head}_{n}(m,\beta_{\ell})\rightarrow 1$ if the event is inferred as head at block $n$ or otherwise  $\mathbb{I}^\text{head}_{n}(m,\beta_{\ell})\rightarrow 0$. 
If the event is not recognized as tail at the last block $N$. The event's label will be considered as head to avoid the false alarm issue. Mathematically,
\begin{equation}
\small
    \mathbb{I}^\text{head}_{N}(m,\beta_\ell,\beta_{\text{u}}) = \mathbb{\sigma}(\beta_\text{u}-C_{N(m)}^\text{tail})\prod\limits_{k=1}^{N-1}\mathbb{\sigma}(\beta_{\text{u}}-C_{k(m)}^\text{tail})\mathbb{\sigma}(C_{k(m)}^\text{tail}-\beta_\ell).
\end{equation}
\subsubsection{Tail Event Detection} Event $m$ is classified as a tail event under the following condition: $C_{n(m)}^\text{tail} > \beta_{\text{u}}, 1\leq n\leq N $.
Before reaching the last block, the condition $\beta_{\ell} \leq C_{n(m)}^\text{tail} \leq\beta_{\text{u}}$ causes the event to continue to the next exit point [i.e., \text{($n$+1)}-th exit]. Similar to \eqref{head}, the conditions can be validated using the following asymptotic indicator function for all $n\in \{1,2...,N\}$:
\begin{equation}\label{tail}
\small
   \mathbb{I}^\text{tail}_{n}(m,\beta_\ell,\beta_{\text{u}}) = \mathbb{\sigma}(C_{n(m)}^\text{tail}-\beta_{\text{u}})\prod\limits_{k=1}^{n-1}\mathbb{\sigma}(\beta_{\text{u}}-C_{k(m)}^\text{tail})\mathbb{\sigma}(C_{k(m)}^\text{tail}-\beta_{\ell}).
\end{equation}
An event is classified locally as either a head or a tail event. We define $\mathbb{L}(x_m,\widehat{x}_m)$ as the loss function such that $\mathbb{L}(x_m,\widehat{x}_m)=0$ if $x_m=\widehat{x}_m$, or otherwise $\mathbb{L}(x_m,\widehat{x}_m)=1$. Based on \eqref{tail}, the indicator function that validates one tail event $m$ is correctly detected using the threshold can be expressed as follows:
\begin{equation}\label{tail correct}
    \mathbb{I}_\text{tail}(x_m,\widehat{x}_m,\beta_\ell,\beta_{\text{u}}) = \sum_{n=1}^{N}\mathbb{I}^\text{tail}_{n}(m,\beta_\ell,\beta_{\text{u}})[1-\mathbb{L}(x_m, \widehat{x}_m)].
\end{equation}
Likewise, a head event $m$ is correctly detected can have the indicator function as follows:
\begin{equation}\label{head correct}
\begin{split}
    \mathbb{I}_\text{head}(x_m,\widehat{x}_m,\beta_{\ell},\beta_\text{u}) = \sum_{n=1}^{N}\mathbb{I}^\text{head}_{n}(m,\beta_{\ell},\beta_\text{u})[1-\mathbb{L}(x_m, \widehat{x}_m)].
\end{split}
\end{equation}
Last, it is worth mentioning that a head event is typically detected at a relatively shallower layer and a tail event at a relatively deeper layer.
\vspace{-4mm}
\subsection{Tradeoff Analysis}
Let $P_{\text{miss}}$ and $P_{\text{false}}$ denote the probability that a tail event is incorrectly detected as a head event and the probability that a head event is falsely detected as a tail event, respectively. Moreover, let $M_\text{tail}$ and $M_\text{head}$ denote the number of tail and head events among a total of $M$ events. Using the notation, we can write:
\begin{equation}\label{miss}
\small
\begin{split}
    P_{\text{miss}}(\beta_\ell,\beta_{\text{u}}) &=  1- \frac{\sum_{m=1}^{M}\mathbb{I}_\text{tail}(x_m,\widehat{x}_m,\beta_\ell,\beta_{\text{u}})}{M} \cdot \frac{M}{M_\text{tail}},\quad M\rightarrow\infty, \\ &= 1- \frac{P_\text{tail, loc}(\beta_\ell,\beta_{\text{u}})}{P_\text{tail}}, 
\end{split}
\end{equation}
where $P_\text{tail, loc}(\beta_\ell,\beta_{\text{u}})$ denotes the probability of events correctly classified as the tail class locally. Likewise, let $P_\text{head}(x,\widehat{x},\beta_{\ell})$ denote the probability that a head event is correctly detected  among all events. From \eqref{head correct}, $P_{\text{false}}$ can be derived as:
 \begin{equation}\label{false}
 \small
 \begin{split}
     P_{\text{false}}(\beta_{\ell},\beta_\text{u})  &= 1- \frac{\sum_{m=1}^{M}\mathbb{I}_\text{head}(x_m,\widehat{x}_m,\beta_{\ell},\beta_\text{u})}{M} \cdot \frac{M}{M_\text{head}}, \quad M\rightarrow\infty, \\ &= 1- \frac{P_\text{head, loc}(\beta_{\ell},\beta_\text{u})}{P_\text{head}},
\end{split}
 \end{equation}
 where $P_\text{head, loc}(\beta_\ell,\beta_{\text{u}})$ denotes the probability of events correctly classified among all events.
We denote $P_{\text{off}}$ as the probability that one event is detected as tail by the dual-threshold mechanism and offloaded to the server, namely the \text{offloading probability}. Utilizing \eqref{miss}, \eqref{false} and considering $M\rightarrow\infty$, we can express $P_{\text{off}}$ using a new form:
\begin{equation}\label{radio}
\begin{split}
   P_{\text{off}}(\beta_{\ell},\beta_{\text{u}}) &= \frac{\sum_{m=1}^{M}\mathbb{I}_\text{tail}(x_m,\widehat{x}_m,\beta_\ell,\beta_{\text{u}})}{M} \\&+ \frac{M_\text{head}-\sum_{m=1}^{M}\mathbb{I}_\text{head}(x_m,\widehat{x}_m,\beta_{\ell},\beta_\text{u})}{M}, M\rightarrow\infty, \\&=   P_\text{tail, loc}(\beta_\ell,\beta_{\text{u}})+P_{\text{head}}-  P_\text{head, loc}(\beta_{\ell},\beta_\text{u}),  \\ &= \left(1-P_{\text{miss}}(\beta_\ell,\beta_{\text{u}})\right)\cdot P_{\text{tail}}+P_{\text{false}}(\beta_{\ell},\beta_\text{u})\cdot P_{\text{head}}.
\end{split}
\end{equation}
Upon identifying an event as tail, the local device offloads the event's relevant features to the server. The server then performs multi-class classification to achieve an accurate recognition of the tail event using a deeper CNN. 

\textbf{“Missing-target-offloading tradeoff”:} It is observed from \eqref{radio} that there exists a fundamental tradeoff between reducing  the missing target probability ($P_\text{miss}$) and reducing the offloading probability ($P_{\text{off}}$). Specifically, a larger  $P_\text{off}$ causes more events to be detected as a tail, which increases the tail event classification accuracy. Although $P_\text{miss}$ decreases, all events must traverse more CNN blocks for detailed classification, which increases the system's communication and computation overhead. The tradeoff is regulated by the dual thresholds ($\beta_\ell$,$\beta_\text{u}$). This necessitates their optimization that is the topic of the next section.

\vspace{-4mm}
\section{Optimal Dynamic Co-inference}\label{DynamicCo-inference:Energy-EfficienteventPartitioning}
 Recall that at the core of dynamic co-inference is the event detection scheme controlled by dual thresholds. In this section, we optimize the thresholds to maximize the E2E co-inference accuracy under constraints on communication overhead (as quantified by the offloading probability) and mobile energy consumption. By showing its convexity, the optimization problem is solved using the gradient descent method. As a result, the optimal offloading policy is revealed to have a threshold based structure that depends on the channel SNR.
 \vspace{-5mm}
\subsection{Problem Formulation}
Given $M$ events and \eqref{tail correct}, the indicator function that validates whether one tail event $m$ is correctly classified at the server can be expressed as\cite{threshold}:
\begin{equation}
\small
    \mathbb{I}_\text{tail}(y_m,\widehat{y}_m,\beta_\ell,\beta_{\text{u}}) = \mathbb{I}_\text{tail}(x_m,\widehat{x}_m,\beta_\ell,\beta_{\text{u}})[1-\mathbb{L}(y_m, \widehat{y}_m)],
\end{equation}
where $y_m$ and $\widehat{y}_m$ correspond to the predicted and ground truth labels at the server, respectively. The tail event's final classification accuracy can then be derived as follows:
\begin{equation}
    \begin{split}
        f_\text{acc}(\beta_\ell,\beta_{\text{u}}) &= \frac{\sum_{m=1}^{M}\mathbb{I}_\text{tail}(y_m,\widehat{y}_m,\beta_\ell,\beta_{\text{u}})}{M} \cdot \frac{M}{M_\text{tail}}, \quad M\rightarrow\infty.
    \end{split}
\end{equation}
Considering the device's average total energy consumption per event detection, denoted as $E_\text{total}$, which comprises two main components: the local computation energy $E_\text{loc}$ and the energy utilized for the offloading of features $E_\text{off}$. This relationship is succinctly captured by the following equation:
\begin{equation}\label{totalenergy}
E_\text{total}(\beta_\ell,\beta_{\text{u}}) =  E_\text{loc}(\beta_\ell,\beta_{\text{u}}) + E_\text{off}(\beta_\ell,\beta_{\text{u}}),
\end{equation}
where $E_\text{loc}$ is determined by summing the local computation energy consumed during all memory access operations required per one event detection. From \eqref{localenergy}, \eqref{head} and \eqref{tail}, this can be mathematically expressed as follows:
\begin{equation}
\small
\begin{split}
    E_\text{loc}(\beta_\ell,\beta_{\text{u}}) &= \frac{\sum_{m=1}^{M}\sum_{n=1}^{N}[ \mathbb{I}^\text{tail}_{n}(m,\beta_\ell,\beta_{\text{u}})+ \mathbb{I}^\text{head}_{n}(m,\beta_\ell,\beta_{\text{u}})]E_\text{loc}(n)}{M}, \\& \quad M \rightarrow\infty.
    \end{split}
\end{equation}
Since the offloading energy is applicable only to events identified as tail events, given the event size $D$, transmit power $P_\text{tr}$ and the data rate $R_\text{tr}$, the offloading energy is formulated as:
\begin{equation}
\begin{split}
     E_\text{off}(\beta_\ell,\beta_{\text{u}}) &= P_{\text{tr}}t_\text{off}, \\& =\frac{P_{\text{tr}}D}{R_\text{tr}M}\sum_{m=1}^{M}\sum_{n=1}^{N}\mathbb{I}^\text{tail}_{n}(m,\beta_\ell,\beta_{\text{u}}), \quad M\rightarrow\infty.
\end{split}
\end{equation}
Given the dynamic nature of channel conditions, which impact offloading energy by altering the data transmission rate, adopting a channel-adaptive optimization strategy is essential. We assume that within each coherence time, the system can detect $M$ events and offload those classified as tail-class to the server. To meet our new energy efficiency goal, we transform the constraint on offloading probability $P_{\text{off}}$ into a constraint on the data transmission volume $v(\beta_{\ell}, \beta_{\text{u}})$. The optimization problem is formulated as follows:
\begin{align}
\textbf{(P1)} \displaystyle \quad \min_{\beta_{\ell},\beta_{\text{u}}} \quad & - f_\text{acc}(\beta_\ell,\beta_{\text{u}}),\\
\textrm{s.t.} \quad & v(\beta_{\ell}, \beta_{\text{u}})  = D\cdot M\cdot P_{\text{off}}(\beta_{\ell},\beta_{\text{u}}) \leq \theta,\\
\quad & f_\text{energy}(\beta_\ell,\beta_{\text{u}}) = M\cdot E_\text{total}(\beta_\ell,\beta_{\text{u}}) \leq \xi.
\end{align}
In \textbf{P1}, the complexity arises from multiple non-convex constraints that intertwine tail event detection accuracy with offloading volume and energy considerations, while also accommodating the variability of communication channels over coherence times. The challenge is to develop a robust mechanism that dynamically adjusts thresholds to maximize the system performance within acceptable data rate and energy limits.
\vspace{-2mm}
\subsection{Channel-Adaptive Threshold Optimization}
\subsubsection{Offloading Feasibility Condition}
Since offloading rare events requires the system to adapt to varying channel conditions, it is essential to understand the channel requirements needed for the system to perform offloading within given energy and data volume constraints. By dynamically adjusting the dual thresholds, the system can classify events at the earliest possible block, minimizing local energy consumption. When all events are detected at the first block of the CNN, local energy use is minimized. If a tail event is detected and offloading is required, the system must ensure that the transmission energy required for offloading remains within the available energy budget: 
$\frac{P_\text{tr}D}{R_\text{tr}}\leq \xi-M\cdot E_\text{loc}(1)$. Under this condition, the offloading feasibility condition is articulated in the following Lemma.
\begin{Lemma}\rm(Feasibility Condition).\label{offloadingcondition}
Given the total number of events $M$, the bandwidth $B$, and the energy constraint $\xi$, the system is feasible to support offloading if the SNR satisfies the following condition:
\begin{equation}\label{lowerbound}
\text{SNR} \ge 2^{\displaystyle \frac{P_\text{tr}D}{B\left(\xi - M \cdot S^{\text{mem}}_{1} \varrho \right)}} - 1.
\end{equation}
\end{Lemma}
The inequality highlights that an increase in data size per event, transmission power, or the total number of events raises the required SNR, as these factors increase the energy demands for data transmission. Conversely, enhanced bandwidth capabilities reduce the required SNR by facilitating quicker data transmission. 

\subsubsection{Optimization of Dual Confidence Thresholds}
To solve \textbf{P1}, it is essential to understand the mathematical properties of both the objective function and constraints, particularly in terms of convexity and smoothness. These properties play a key role in determining the behavior of optimization algorithms and their convergence. To this end, we analyze and demonstrate the Lipschitz continuity of the functions in \textbf{P1}.
\begin{Lemma}\label{gradientcontinuousobject}\rm
    The gradient of objective function $f_{\text{acc}}(\beta_\ell,\beta_\text{u})$ is Lipschitz continuous with the Lipschitz constant $\gamma= k^2\frac{N(N+1)(N+4\sqrt{3}-1)}{24}$. \\
\end{Lemma}
\vspace{-3mm}
\noindent The proof is given in Appendix \ref{lemma2}.

\begin{Lemma}\label{energycontinuity}\rm
   $v(\beta_{\ell}, \beta_{\text{u}})$'s gradient is Lipschitz continuous with the Lipschitz constant $2DM\gamma$. $f_\text{energy}(\beta_\ell, \beta_\text{u})$'s gradient is Lipschitz continuous with the Lipschitz constant $2M\gamma\left(E_\text{loc}(N) + \frac{P_{\text{tr}}D}{2R_\text{tr}}\right)$. \\
\end{Lemma}
\vspace{-3mm}
\noindent The proof is given in Appendix \ref{lemma3}.

Since all functions are proved to have a Lipschitz continuous gradient, the convexity and smoothness of each function in \textbf{P1} are analyzed in the following Lemma.
\begin{Lemma}\rm \label{convexandsmooth}
     $f_{\text{acc}}(\beta_\ell,\beta_\text{u})$ is a $\gamma$-weakly convex and $\gamma$-smooth function. $v(\beta_{\ell}, \beta_{\text{u}})$ is a $2DM\gamma$-weakly convex and $2DM\gamma$-smooth function. Given the data rate $R_\text{tr}$, transmit power $P_{\text{tr}}$, and event data size $D$, $f_\text{energy}(\beta_\ell, \beta_\text{u})$ exhibits $2M\gamma\left(E_\text{loc}(N) + \frac{P_{\text{tr}}D}{2R_\text{tr}}\right)$ weak convexity and smoothness. \\
\end{Lemma}
\vspace{-3mm}
\noindent The proof is given in Appendix \ref{proposition2}.

Subsequently, we propose an approach to obtain an efficient solution to problem \textbf{P1} using the proximal-point penalty method \cite{lin2022complexity}. This method involves constructing a proximal term and a quadratic penalty term, which associate the constraint function with the original objective function. For $t$-th iteration, it performs the update:
\begin{equation}
    (\overline{\beta}_\ell^{t+1}, \overline{\beta}_\text{u}^{t+1}) = \arg\min_{\beta_\ell,\beta_\text{u}}f_{t}(\beta_\ell,\beta_\text{u}),
\end{equation}
where $f_{t}(\beta_\ell,\beta_\text{u})$ denotes the proximal penalty function. Mathematically,
\vspace{-3mm}
\begin{equation}\label{energyippp}
\begin{split}
  {f}_{t}(\beta_\ell,\beta_\text{u}) &= f_\text{acc}(\beta_\ell,\beta_{\text{u}}) + \frac{\lambda}{2}\left\Vert\begin{pmatrix}
\beta_\ell-\overline{\beta}_\ell^{t} \\
   \beta_\text{u}-\overline{\beta}_\text{u}^{t}
\end{pmatrix}\right\Vert^2 \\&\quad+\frac{\kappa}{2}\left(\max \{ 0,v(\beta_{\ell}, \beta_{\text{u}})\}\right)^2 \\&\quad+ \frac{\rho}{2}\left(\max \{ 0,f_\text{energy}(\beta_\ell,\beta_\text{u})\}\right)^2.
\end{split}
\end{equation}
Leveraging the Lemma \ref{convexandsmooth}, the convexity and smoothness of the function ${f}_{t}(\beta_\ell, \beta_{\text{u}})$ are established in the Proposition \ref{energyipppconvexsmooth}. 
\begin{algorithm}[t]
\caption{Channel Adaptive Dual-threshold Optimization}
\label{alg:dynamic dual_threshold}
\begin{algorithmic}[1] 
    \STATE \textbf{Input:} Proximal parameter $\lambda$, Penalty parameters $\kappa$, $\rho$, Smoothness parameter $\psi$, Convexity parameter $\eta$, Transmit power $P_\text{tr}$, Bandwidth $B$, Data size $D$, Event number $M$, The local energy consumption at the first block $S^{\text{mem}}_{1} \varrho$.
    \STATE \textbf{Initialize:} $\overline{\beta}_\text{prox}^0 = \begin{pmatrix} \overline{\beta}_\ell^0 \\ \overline{\beta}_\text{u}^0 \end{pmatrix}$, $\overline{\beta}_\text{extra}^0 = \begin{pmatrix} \overline{\beta}_\ell^0 \\ \overline{\beta}_\text{u}^0 \end{pmatrix}$ 

    \FOR{$c = 1$ to $ C$} 
        \IF{$\text{SNR} \ge 2^{\displaystyle \frac{P_\text{tr}D}{B\left(\xi - M \cdot S^{\text{mem}}_{1} \varrho \right)}} - 1$}
            \STATE Update $R_\text{tr}(c)$, $\psi_c$ and $\eta_c$ based on current channel conditions
            \STATE Update $E_\text{off}$ and $f^c_\text{energy}$ based on the new $R_\text{tr}(c)$
            
            \FOR{$t = 0$ to $T-1$}
            \STATE Define $f_t(\overline{\beta}) = f_{\text{miss}}(\overline{\beta}) + \frac{\lambda}{2} \|\overline{\beta} - \overline{\beta}^t\|^2 + \frac{\kappa}{2} (\max \{0, P_{\text{off}}(\overline{\beta})\})^2+\frac{\rho}{2}\left(\max \{ 0,f^c_\text{energy}(\overline{\beta})\}\right)^2$
            
                \FOR{$i = 0$ to $I-1$}
                \STATE $\overline{\beta}_\text{prox}^{(i+1)} = \text{Prox}_{\lambda, \kappa}\left(\overline{\beta}_\text{extra}^{(i)} - \frac{1}{\psi_c} \nabla f_t(\overline{\beta}_\text{extra}^{(i)})\right)$ //Proximal operator
                \STATE $\overline{\beta}_\text{extra}^{(i+1)} = \overline{\beta}_\text{prox}^{(i+1)} + \frac{\sqrt{\psi_c} - \sqrt{\eta_c}}{\sqrt{\psi_c} + \sqrt{\eta_c}} (\overline{\beta}_\text{prox}^{(i+1)} - \overline{\beta}_\text{prox}^{(i)})$ //Extrapolation step
                \ENDFOR
                \STATE $\overline{\beta}^{t+1} = \overline{\beta}_\text{prox}^{(I)}$
            \ENDFOR
        \ENDIF
\STATE \textbf{Output:} $\overline{\beta}$ that minimizes $\| \overline{\beta}^{t+1} - \overline{\beta}^t \|$ across all $t$
\ENDFOR
\end{algorithmic}
\end{algorithm}

\begin{Proposition}\label{energyipppconvexsmooth}\rm
Given a sufficiently large proximal parameter \(\lambda\), the function ${f}_{t}(\beta_\ell, \beta_{\text{u}})$ is strongly convex, with its smoothness parameter $\psi$ and strong convexity parameter $\eta$ given by:
\begin{equation}
\small
\begin{split}\label{energyfunctionsmooth}
&\psi = \gamma + \lambda + \kappa DMA\left(A + 2\gamma\right) \\&\quad+ \rho B\left(B + 2M\gamma \left(E_\text{loc}(N) + \frac{P_{\text{tr}}D}{2B\log\left(1+\text{SNR}\right)}\right)\right),
\end{split}
\end{equation}
\begin{equation}
\small
\begin{split}\label{energyfunctionconvex}
\eta &= \lambda - \gamma - 2M\gamma \left(\kappa AD + \rho B\left(E_\text{loc}(N) + \frac{P_{\text{tr}}D}{2B\log\left(1+\text{SNR}\right)}\right)\right),
\end{split}
\end{equation}
where $A$ and $B$ are constants determined by the condition:
\begin{equation}
\small
    A = \max\left\{\theta,\frac{DM(N-1)}{2\sqrt{2}}\right\},
\end{equation}
\begin{equation}
\small
B = \max\left\{\xi, \frac{(N^2 + 1)E_\text{loc}(N)}{2\sqrt{2}} + \frac{(N + 2)(N - 1)P_{\text{tr}}D}{4\sqrt{2}B\log\left(1+\text{SNR}\right)}\right\}.
\end{equation}
\end{Proposition}
\noindent The proof is given in Appendix \ref{Proposition3}.

Acknowledging the smoothness and strong convexity parameter of ${f}_{t}(\beta_\ell, \beta_{\text{u}})$, an algorithm based on accelerated gradient descent is developed to update the optimal dual thresholds. After the offloading feasibility condition is checked using \eqref{lowerbound}, the algorithm adjusts the offloading decision and data transmission rate for each possible channel condition. By applying the proximal gradient method, the algorithm refines the thresholds to maximize event classification accuracy while maintaining energy efficiency. Additionally, it precomputes optimal thresholds for various channel conditions and constraints, storing them in a lookup table. During operation, the algorithm quickly references these thresholds based on the current SNR, ensuring efficient and accurate decision-making. The operational mechanics are detailed in Algorithm 1. As the smoothness and strong convexity parameter of the function ${f}_{t}(\beta_\ell,\beta_\text{u})$ depend on channel SNR, the algorithm’s efficiency is inherently influenced by the channel quality, which will be discussed in the following Remark.

\begin{Remark}\rm
    With fixed transmit power $P_\text{tr}$ and event data size $D$, the convergence rate for resolving each sub-problem in Algorithm 1 is expressed as:
    \begin{equation}
        {f}(\overline{\beta}^{(i)}) - {f}_t(\overline{\beta}^{*}) \leq {O}\left(\left(1-\sqrt{\frac{\eta}{\psi}}\right)^{i}\right).
    \end{equation}
   The convergence rate of this dynamic algorithm is modulated by channel conditions. Specifically, enhanced channel gain increases $R_\text{tr}$, which in turn accelerates the convergence process, optimizing the algorithm's performance in variable network environments. Conversely, an increase in the number of events, transmission power, or event size can adversely affect system efficiency by expanding the computational and communication demands for each iteration. 
\end{Remark}
\subsubsection{Structure of Optimal Policy}
With the optimal dual thresholds identified through our adaptive optimization framework, the system generates a responsive optimal offloading policy that aligns with real-time channel conditions to ensure resource-efficient operation. The following proposition formalizes this offloading policy, which dynamically adjusts the number of events processed locally versus offloaded to the server according to the varied SNR. The system's channel-adaptive offloading policy can be formulated in the Proposition \ref{offloadingpolicy}.
\begin{Proposition}\rm (Threshold based Offloading Policy)\label{offloadingpolicy}.
Given $\left(\beta_\ell^*,\beta_\text{u}^*\right)$ is the optimal solution stored in the lookup table for the classification of $M$ events under the data volume constraint $\theta$, energy constraint $\xi$, and the channel condition specified by the SNR, the co-inference system's optimal offloading policy has the following structure:
\begin{enumerate}
    \item Given each event's size is $D$, none of these events are offloaded if the SNR does not meet the feasibility condition in (\ref{lowerbound}), i.e.,
    \begin{equation}
        M^*_\text{off}=0, \quad \text{SNR} < 2^{\displaystyle \frac{P_\text{tr}D}{B\left(\xi - M \cdot S^{\text{mem}}_{1} \varrho \right)}} - 1.
    \end{equation}
    \item The system offloads $M_\text{off}$ events based on different channel conditions, specifically:
\begin{equation}
\small
    \begin{split}
         &M^*_\text{off}\displaystyle = \Bigg\lfloor\frac{B\left(\xi - M\cdot E_\text{loc}\left(\beta_\ell^*,\beta_\text{u}^*\right)\right)\log(1+\text{SNR})}{P_\text{tr}D}\Bigg\rfloor, \\&\quad \text{SNR} \ge 2^{\displaystyle \frac{P_\text{tr}D}{B\left(\xi - M \cdot S^{\text{mem}}_{1} \varrho \right)}} - 1.
    \end{split}
\end{equation}

\end{enumerate}
\end{Proposition}

The Proposition highlights that the adaptive offloading policy strategically dynamically adjusts the dual thresholds based on SNR conditions, balancing classification accuracy and resource efficiency under a fixed energy constraint. This adaptive mechanism optimally selects thresholds for event classification by adjusting to different SNR scenarios: 
\begin{enumerate}
    \item \textbf{High SNR:} With increased transmission rates under the same energy constraint, the energy cost for offloading each event decreases, enabling the system to adjust the dual thresholds to offload more events with greater confidence. The system has two possible strategies for adjusting the dual thresholds. The first strategy is to lower the upper threshold $\beta_\text{u}$, allowing more events to be classified as tail and subsequently offloaded to the server, effectively reducing the miss rate for tail events. Alternatively, the system can expand the uncertainty area by lowering the lower threshold $\beta_\ell$ and/or increasing the $\beta_\text{u}$, permitting more events with less certain confidence scores to pass through additional local CNN layers. This adjustment ensures that events are more accurately analyzed locally before offloading, leading to improved data reliability upon reaching the server and enhanced overall classification accuracy.
    \item \textbf{Low SNR:} Under low SNR conditions, where transmission constraints restrict data offloading, the system adjusts the dual thresholds by lowering $\beta_\text{u}$ and raising $\beta_\ell$, effectively narrowing the uncertainty area. This adjustment prioritizes earlier classification of head events locally, helping to maintain a manageable false alarm rate while allowing the system to conserve computation energy by quickly filtering out head events without unnecessary processing.
\end{enumerate}

\section{Experimental Result}\label{ExperimentResult}
\subsection{Experimental Settings}
The default experimental settings are as follows unless specified otherwise.
\begin{itemize}
\item \textbf{DNN Model and Dataset Setting:}
In this study, the ShuffleNetV2 \cite{ma2018shufflenet} and MobileNetV2 \cite{sandler2018mobilenetv2} models are deployed on the local device, with each block followed by an intermediate classifier. The ResNet50\cite{resnet} model is deployed on the server. The local models were pretrained on a well-known dataset\cite{naturedataset} comprising 25,000 retinal images, including both normal and unhealthy retina images, for binary classification. The server model (ResNet50) was pretrained on the same dataset with one normal class and three unhealthy retina classes for multi-class classification. 
To explore the effects of class imbalance, each local CNN model was further trained using two separate datasets with imbalance ratios of 4:1 and 9:1. Here, we consider each image to be one event. For validation and testing, distinct sets of 1,250 images are used. The entire validation set was utilized to determine the optimal threshold for classification. Subsequently, the 1,250 test images were divided into five groups of 250 images each, maintaining the same class imbalance ratio. Independent classification tests were conducted on each group using the optimal threshold. Finally, the overall model performance was assessed by averaging the classification accuracies and missing probabilities across the five groups.

    \item \textbf{Communication Settings:}
    We consider $M$ images sequentially detected locally and decide whether to offload them to the server for tail classification. To reduce the communication overhead, each image is resized from 3 × 224 × 224 to 3 × 56 × 56 (i.e., reducing the image resolution) for offloading. Assuming perfect channel state information (CSI) at the server, the transmission power $P_\text{tr}$ is fixed at 30 dBm, and the bandwidth $B$ is set to 30 MHz. The fading coefficient $h$ is varied to control the SNR, calculated as $\text{SNR} = \frac{|h|^2 P_\text{tr}}{P_n}$ where $P_n$ is the noise power. This setup allows us to observe the impact of different SNR levels on the data rate, computed using Shannon's theorem: $R_\text{tr} = B \log_2(1 + \text{SNR})$. By adjusting \(h\), we simulate varying channel conditions to study their effect on communication performance.

\end{itemize}

Three benchmarking schemes from the standard inference model are adopted, as described below.
\begin{itemize}
    \item \textbf{Single Threshold Detection}:
    This traditional approach\cite{exiting} employs intermediate classifiers in conjunction with a single threshold mechanism throughout the network. At each intermediate stage, the classification decision is based on the highest confidence score for a class, provided this score exceeds a predefined threshold. If, by the final block, no class's confidence score surpasses this threshold, the event is automatically assigned to the head class.
    \item \textbf{Terminal Detection}:
   This method eschews intermediate classifiers and implements a single decision threshold solely at the final stage of the network\cite{terminaldetection}. Every event undergoes full processing through the entire network, culminating in a single classification decision at the end. The event is classified as the tail class if the Softmax-processed confidence level for the tail class at the final classifier exceeds a specific threshold. 
    \item \textbf{Ideal Case}:
     In an ideal tail-detection system, it is assumed that all event labels are accurately identified at the first block on the local device without any errors. This implies that the system flawlessly distinguishes between head and tail classes, ensuring that no instances of head classes are incorrectly classified as tail classes.
\end{itemize}
\begin{figure}[t!]
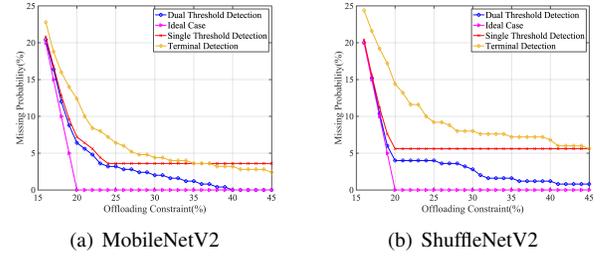

\centering
\subfigure[MobileNetV2]{
\includegraphics[width=0.45\columnwidth]{figures/MobileNetV2_miss_off.pdf}\label{MobileNetV2_miss_off}}
\subfigure[ShuffleNetV2]{
\includegraphics[width=0.45\columnwidth]{figures/ShuffleNetV2_miss_off.pdf}\label{ShuffleNetV2_miss_off}}
\caption{Missing probability versus offloading constraints in the case of imbalanced ratio $R=4$.}
\label{fig:miss_offload}
\vspace{-5mm}
\end{figure}

\subsection{Comparison of Detection Mechanism with CNN Models}
In our comparative analysis of threshold-based classification strategies using CNNs, we varied the offloading constraint from 16\% to 45\% in 1\% intervals to evaluate the effectiveness of our proposed dual threshold detection mechanism against established benchmarks. Using datasets with an imbalance ratio of 4:1, we assessed performance on the MobileNetV2 and ShuffleNetV2 architectures, as shown in Fig. \ref{MobileNetV2_miss_off} and \ref{ShuffleNetV2_miss_off}. The dual threshold mechanism outperformed the other methods in identifying the minority class by employing two flexible confidence thresholds for early decision-making, thereby reducing misclassification and overfitting by intermediate classifiers. In both models, the performance gain of purposed dual threshold detection schemes increases because the terminal detection fails to accurately detect more complex events as the offloading constraint increases. The single threshold detection initially showed an advantage over terminal detection due to its early-stopping capability, which prevents overfitting and enhances generalization. However, the benefits of the single threshold mechanism plateaued as the offloading probability continued to rise. This is because its minimum threshold value of 0.5 led to consistent misclassification of low-confidence events as belonging to the head class. Compare to the MobileNetV2, dual threshold detection demonstrated superior performance in ShuffleNetV2 at low offloading constraint, due to ShuffleNetV2's efficient architecture featuring group convolutions and channel shuffling, which allows it to quickly adapt and perform well with limited input. Conversely, as offloading rates and event complexity increased, MobileNetV2's deeper network architecture, with depthwise separable convolutions, exhibited superior feature extraction and class discrimination.
\begin{figure}[t!]
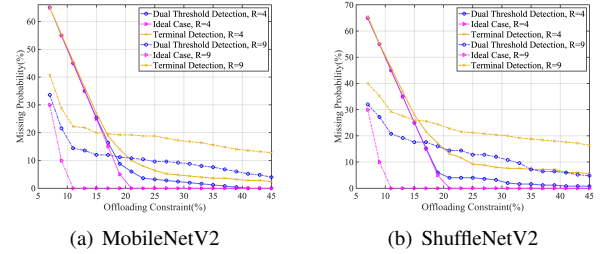

\centering
\subfigure[MobileNetV2]{
\includegraphics[width=0.45\columnwidth]{figures/MobileNetV2_imbalance_ratio.pdf}\label{MobileNetV2_imbalance_ratio}}
\subfigure[ShuffleNetV2]{
\includegraphics[width=0.45\columnwidth]{figures/ShuffleNetV2_imbalance_ratio.pdf}\label{ShuffleNetV2_imbalance_ratio}}
\caption{Missing probability versus offloading constraints in the cases of imbalanced ratio $R=4$ and $R=9$.}
\label{fig:miss_offload}
\vspace{-3mm}
\end{figure}

\vspace{-4mm}
\subsection{Detection Mechanism Analysis across Imbalanced Datasets}
We continue our investigation into the efficacy of different detection mechanisms by analyzing their performance across datasets that exhibit varying degrees of class imbalance, as demonstrated in Fig.\ref{MobileNetV2_imbalance_ratio} and \ref{ShuffleNetV2_imbalance_ratio}. The datasets are bifurcated into two categories with imbalanced ratios of 4:1 and 9:1, whilst maintaining an equivalent total number of events.The single threshold approach has been excluded from this analysis because it rapidly reaches the offloading limit in highly imbalanced datasets. In such datasets, many tail events have confidence scores below 0.5, which prevents their offloading and limits the approach's effectiveness.
As offloading probabilities increase, the dual threshold method consistently outperforms the terminal detection scheme, particularly in the dataset ratio $R=9$. This suggests that as the offloading probability grows, each system begins to struggle with more challenging instances of the tail class that are not as easily separable from the head class. The increased imbalance ratio exacerbates this challenge, as the prevalence of the majority class (head class) introduces bias in the classifiers. The slower reduction in the miss rate indicates a trade-off indicates a trade-off where improved tail class detections come at the cost of more head class misclassifications. This performance gain highlights the dual threshold method's robustness in handling highly imbalanced data scenarios and its effectiveness in maintaining classification accuracy under varying offloading constraints.

\begin{figure}[t!]
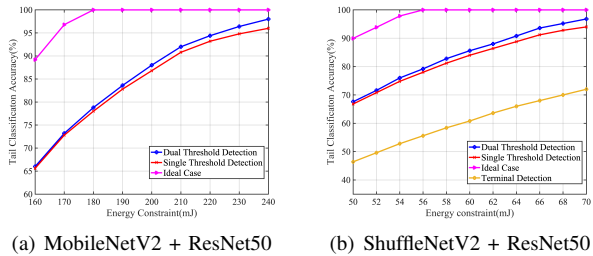

\centering
\subfigure[MobileNetV2 + ResNet50]{
\includegraphics[width=0.45\columnwidth]{figures/MobileNetV2_energy_offload.pdf}\label{MobileNetV2_energy}}
\subfigure[ShuffleNetV2 + ResNet50]{
\includegraphics[width=0.45\columnwidth]{figures/ShuffleNetV2_energy_offload.pdf}\label{ShuffleNetV2_energy}}
\caption{Tail classification accuracy versus energy constraints in the case of imbalanced ratio $R=4$, SNR = 5dBm. }
\label{fig:miss_offload}
\vspace{-3mm}
\end{figure}
\subsection{Detection Mechanism Analysis across Energy Constraint}
We proceeded to analyze the performance of various detection models under different energy constraints, focusing on the efficiency of early-stopping mechanisms within single and dual threshold models. The offloading data volume constraint was set at 0.7 MB, with an SNR of 5 dB. The simulation results are illustrated in Fig. \ref{MobileNetV2_energy} and \ref{ShuffleNetV2_energy}. The ideal curve assumes that all events are detected at the first block, allowing the remaining energy to be utilized for offloading all tail events to the server for classification.
The terminal threshold detection, lacking an early-stopping feature, cannot be fairly compared to the other two schemes on MobileNetV2 under the same energy constraints due to MobileNetV2's significant depth, which incurs substantial computational energy consumption. This approach requires traversing all network blocks to reach a classification decision. Although it achieves classification accuracy on ShuffleNetV2, it exhibits the worst performance in terms of energy efficiency compared to the other schemes, owing to its inherently higher energy consumption.
Under identical energy constraints, the dual threshold model consistently demonstrates higher classification accuracy than the single threshold and terminal threshold schemes. The performance gain becomes more significant under increasing energy constraints because the dual threshold model can flexibly adjust its two thresholds to reduce the misclassification of challenging head-class events as tail-class, thereby conserving energy. However, as event complexity increases, accuracy converges due to the need for more network blocks or incorrect tail-class labeling, both increasing energy consumption.
Additionally, the ShuffleNetV2 model demonstrates lower energy consumption and a smaller discrepancy between the real and ideal cases compared to MobileNetV2. Consequently, while deeper networks can achieve higher maximum accuracy, they present significant challenges for energy-constrained applications. This underscores the necessity of balancing complexity and energy efficiency in the design of CNNs.
\begin{figure}[t!]
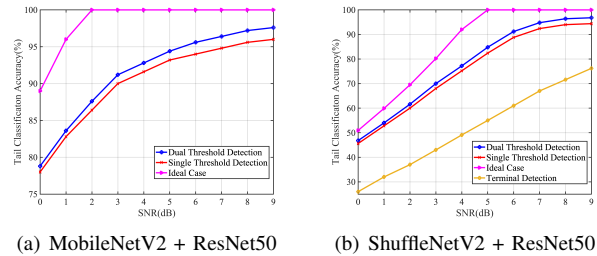

\centering
\subfigure[MobileNetV2 + ResNet50]{
\includegraphics[width=0.45\columnwidth]{figures/MobileNet_snr_miss.pdf}\label{mobilenetV2_snr}}
\subfigure[ShuffleNetV2 + ResNet50]{
\includegraphics[width=0.45\columnwidth]{figures/ShuffleNetV2_snr_miss.pdf}\label{SHuffleNetV2_snr}}
\caption{Tail classification accuracy versus SNR in the case of imbalanced ratio $R=4$, data volume constraint = 0.7MB.}
\label{fig:miss_offload}
\vspace{-3mm}
\end{figure}
\vspace{-4mm}
\subsection{Detection Performance Analysis with Varied Channel Conditions}
In this study, we investigate the system's performance across various channel states. Our comparative analysis involved subjecting the ShuffleNetV2+ResNet50 model to a 210mJ energy constraint and the MobileNetV2 to a 60mJ limit, both with an offloading data volume constraint of 0.7MB, against a dataset with a 4:1 imbalanced ratio. The corresponding simulation results are presented in Fig.\ref{mobilenetV2_snr} and Fig.\ref{SHuffleNetV2_snr}. As the SNR increases, there is a concomitant performance gain in tail classification accuracy for dual threshold detection mechanism. This relationship can be attributed to enhanced transmission rates at higher SNRs, which reduce the energy cost for event offloading. Consequently, within a fixed energy budget, improved transmission efficiency effectively liberates additional energy, which can be reallocated to processing events through more network blocks for precise detection.
The widening gap between schemes and the ideal scenario at higher SNRs indicates that more complex tail-class events require traversal through additional network blocks for correct identification, thereby consuming more energy. The single threshold detection model, despite improvements in SNR, fails to match the performance of the dual threshold model. This is because the single threshold model tends to overfit and lacks an early exiting mechanism, making it less effective at distinguishing between head and tail classes in an imbalanced dataset.
At high SNRs, the tail classification accuracy converges due to the offloading data volume constraint. Within this constraint, the system can offload a maximum of approximately 30\% of events within a given period, achieving the corresponding minimum missing probability, as demonstrated in previous Fig. \ref{MobileNetV2_miss_off} and \ref{ShuffleNetV2_miss_off}.

\section{Conclusion}\label{Conclusion}
In this paper, we have proposed a dynamic co-inference framework designed to alleviate communication and computation bottlenecks while preserving high inference accuracy for long-tail distributed events. Our system enables local binary tail detection and optimizes thresholds for offloading tail events to the server for multi-class classification, balancing the event's missing target probability and offloading probability. To achieve high detection accuracy, we developed a dual-threshold classifier within early-exiting models. Additionally, to optimize E2E tail event classification at the server while considering the device's energy limitations, we designed an offloading policy and presented an online algorithm for determining the optimal confidence threshold for feature offloading. The optimization challenge was addressed by transforming non-convex functions into a strongly convex problem. 

This work is the first to explore event-triggered cooperative inference systems with a novel dual-threshold based architecture design, opening numerous opportunities for future research. One direction could focus on implementing the architecture in distributed sensing for multi-view systems, necessitating further exploration of feature dimensionality's impact on offloading priority. Additionally, integrating more stringent transmission requirements, such as ultra-reliable low-latency communication systems in the context of 6G, warrants further investigation.

\appendices
\vspace{1mm}\section{Proof of Lemma \ref{gradientcontinuousobject}}\label{lemma2}
Consider the single event scenario when $m = m_1$. $\forall n\in{1,2,...,N-1} $ and let $g_n^\text{tail}({\beta_\ell,\beta_\text{u}}) = \mathbb{I}^\text{tail}_{n}(m_1,\beta_\ell,\beta_{\text{u}})$, $g_n^\text{head}({\beta_\ell,\beta_\text{u}}) = \mathbb{I}^{\text{head}}_{n}(m_1,\beta_{\ell},\beta_\text{u})$ and $g_N^\text{tail}({\beta_\ell,\beta_\text{u}}) = \mathbb{I}^\text{tail}_{N}(m_1,\beta_\ell,\beta_{\text{u}})$ . Since $|\sigma(x)|\leq1, |\sigma(x)^{'}|\leq\frac{1}{4}, |\sigma(x)^{''}|\leq\frac{\sqrt{3}}{6}$, an upper bound of $g_n^\text{tail}({\beta_\ell,\beta_\text{u}})$'s spectral norm can be derived as:
\begin{equation}
\small
    \Vert\nabla ^{2}g_n^\text{tail}({\beta_\ell,\beta_\text{u}})\Vert_2\leq\Vert\nabla ^{2}g_n^\text{tail}({\beta_\ell,\beta_\text{u}})\Vert_\text{F}\leq k^2(\frac{n^2-n}{8}+\frac{\sqrt{3}n}{3}).
\end{equation}
Indeed, this upper bound also applies to $g_n^\text{head}({\beta_\ell,\beta_\text{u}})$ and $\mathbb{I}^\text{tail}_{N}(m,\beta_\ell,\beta_{\text{u}})$, as they share the same structure of consecutive Sigmoid function products as $P_n(\beta_\text{u})$. Likewise, it is straightforward that
\begin{equation}
\small
    \Vert\nabla ^{2}g_n^\text{head}({\beta_\ell,\beta_\text{u}})\Vert_2\leq\Vert\nabla ^{2}g_n^\text{head}({\beta_\ell,\beta_\text{u}})\Vert_\text{F}\leq k^2(\frac{n^2-n}{8}+\frac{\sqrt{3}n}{3}),
\end{equation}
\begin{equation}
\small
     \Vert\nabla^{2}g_N^\text{tail}({\beta_\ell,\beta_\text{u}})\Vert_2\leq  k^2(\frac{N^2-N}{8}+\frac{\sqrt{3}N}{3}).
\end{equation}
Since $g_n^\text{tail}({\beta_\ell,\beta_\text{u}})$, $g_n^\text{head}({\beta_\ell,\beta_\text{u}})$ and $\mathbb{I}^\text{tail}_{N}(m,\beta_\ell,\beta_{\text{u}})$ all have hessians with bounded norms, they all follow Lipschitz continuous gradient. Let $\bm\beta = \begin{pmatrix}
\beta_\ell-\overline{\beta}_\ell \\
   \beta_\text{u}-\overline{\beta}_\text{u}
\end{pmatrix}$ denotes the differential vector between two threshold pairs $\left(\beta_\ell,\beta_\text{u}\right)$ and $\left(\overline{\beta}_\ell,\overline{\beta}_\text{u}\right)$, the inequality holds for that:
\begin{equation}\label{inequality1}
\small
   \left\Vert\nabla g_n^\text{tail}({\beta_\ell,\beta_\text{u}}) - \nabla g_n^\text{tail}({\overline{\beta}_\ell,\overline{\beta}_\text{u}})\right\Vert\leq k^2(\frac{n^2-n}{8}+\frac{\sqrt{3}n}{3})\cdot\left\Vert\bm\beta\right\Vert,
\end{equation}
\begin{equation}\label{inequality2}
\small
     \left\Vert\nabla g_n^\text{head}({\beta_\ell,\beta_\text{u}}) - \nabla g_n^\text{head}({\overline{\beta}_\ell,\overline{\beta}_\text{u}})\right\Vert\leq k^2(\frac{n^2-n}{8}+\frac{\sqrt{3}n}{3})\cdot\left\Vert\bm\beta\right\Vert,
\end{equation}
\begin{equation}\label{inequality3}
\small
    \left\Vert\nabla g_N^\text{tail}({\beta_\ell,\beta_\text{u}}) - \nabla g_N^\text{tail}({\overline{\beta}_\ell,\overline{\beta}_\text{u}})\right\Vert\leq k^2(\frac{N^2-N}{8}+\frac{\sqrt{3}N}{3})\cdot\left\Vert\bm\beta\right\Vert.
\end{equation}
Since $\mathbb{I}_\text{tail}(x_m,\widehat{x}_m,\beta_\ell, \beta_\text{u})$ contains $g_n^\text{tail}({\beta_\ell,\beta_\text{u}})$, deriving the partial derivative of $\mathbb{I}_\text{tail}(x_m,\widehat{x}_m,\beta_\ell, \beta_\text{u})$ is equalized to calculating the gradient of $g_n^\text{tail}({\beta_\ell,\beta_\text{u}})$. Mathematically, 
\begin{equation}\label{objective Lipschitz}
\small
\begin{split}
    &\left\Vert \frac{\partial \mathbb{I}_\text{tail}(y_m,\widehat{y}_m,\beta_\ell, \beta_\text{u})}{\partial\beta_\ell\partial \beta_\text{u}} - \frac{\partial \mathbb{I}_\text{tail}(y_m,\widehat{y}_m,\overline{\beta}_\ell,\overline{\beta}_\text{u})}{\partial\overline{\beta}_\ell\partial \overline{\beta}_\text{u}}\right\Vert \\&\overset{(a)}{\leq} \left\Vert\sum_{n=1}^{N}\left(\frac{g_n^\text{tail}({\beta_\ell,\beta_\text{u}})}{\partial\beta_\ell\partial \beta_\text{u}} - \frac{g_n^\text{tail}({\overline{\beta}_\ell,\overline{\beta}_\text{u}})}{\partial\overline{\beta}_\ell\partial \overline{\beta}_\text{u}}\right)\right\Vert \\&\overset{(b)}{\leq}
    k^2\frac{N(N+1)(N+4\sqrt{3}-1)}{24}\cdot\left\Vert\bm\beta\right\Vert \\&= \gamma\cdot\left\Vert\bm\beta\right\Vert,
\end{split}
\end{equation}
where $\gamma= k^2\frac{N(N+1)(N+4\sqrt{3}-1)}{24}$ denotes the Lipschitz constant of the gradient of $\mathbb{I}_\text{tail}(x_m,\widehat{x}_m,\beta_\ell, \beta_\text{u})$, $(a)$ arises from the fact that $1-\mathbb{L}(x_m, \widehat{x}_m)\leq 1$; $1-\mathbb{L}(y_m, \widehat{y}_m)\leq 1$.  $(b)$ arises from the relation that
\begin{equation}
\small
\begin{split}
   &\sum_{n=1}^{N}\left\Vert \nabla g_n^\text{tail}({\beta_\ell,\beta_\text{u}}) - \nabla g_n^\text{tail}({\overline{\beta}_\ell,\overline{\beta}_\text{u}}) \right\Vert \\&\leq k^2\left[\sum_{n=1}^{N}\left(\frac{n^2-n}{8}+\frac{\sqrt{3}n}{3}\right)\right]\cdot\left\Vert\bm\beta\right\Vert.
\end{split}
\end{equation}
Since the Lipschitz constants in \eqref{objective Lipschitz} does not depend on specific events, we extend and generalize the preceding inequality into our objective function:
\begin{equation}\label{}
  \left\Vert\nabla f_{\text{acc}}(\beta_\ell,\beta_\text{u}) -\nabla f_{\text{acc}}(\overline{\beta}_\ell,\overline{\beta}_\text{u}) \right\Vert\leq \gamma\cdot\left\Vert\bm\beta\right\Vert.
\end{equation}
Therefore, the gradient of objective function $f_{\text{acc}}(\beta_\ell,\beta_\text{u})$ is proved to be Lipschitz continuous with the constant $\gamma= k^2\frac{N(N+1)(N+4\sqrt{3}-1)}{24}$.
\noindent Proof of Lemma \ref{lemma2} is completed.

\vspace{-2mm}
\section{Proof of Lemma \ref{energycontinuity}}\label{lemma3}
Let $h(x_m,\widehat{x}_m,\beta_{\ell},\beta_{\text{u}})$ denote a sequence of function $P_{\text{off}}(\beta_\ell,\beta_\text{u})$'s nominator, which $h(x_m,\widehat{x}_m,\beta_{\ell},\beta_{\text{u}}) = \mathbb{I}_\text{tail}(x_m,\widehat{x}_m,\beta_\ell,\beta_\text{u}) -\mathbb{I}_\text{head}\left((x_m,\widehat{x}_m,\beta_{\ell},\beta_\text{u}\right)+M_\text{head}$. According to \eqref{inequality1}, \eqref{inequality2} and \eqref{inequality3}, it follows that
\begin{equation}\label{constraint lipschitz}
\small
\begin{split}
    &\left\Vert \frac{\partial h(x_m,\widehat{x}_m,\beta_{\ell},\beta_{\text{u}})}{\partial\beta_\ell\partial \beta_\text{u}} - \frac{\partial h(x_m,\widehat{x}_m,\overline{\beta}_{\ell},\overline{\beta}_{\text{u}})}{\partial\overline{\beta}_\ell\partial \overline{\beta}_\text{u}}\right\Vert \\&= \left\Vert   \sum_{n=1}^{N}B_n\right\Vert[1-\mathbb{L}(x_m, \widehat{x}_m)]      \\&\overset{(c)}{\leq} \sum_{n=1}^{N}2k^2(\frac{n^2-n}{8}+\frac{\sqrt{3}n}{3})\cdot \left\Vert \bm\beta\right\Vert\\ &= k^2\frac{N(N+1)(N+4\sqrt{3}-1)}{12}\cdot\left\Vert\bm\beta\right\Vert \\&= 2\gamma\cdot\left\Vert \bm\beta\right\Vert.
\end{split}
\end{equation}
Here, $B_n$ denote the summation of gradient differences from \eqref{inequality1} and  \eqref{inequality2}. Mathematically,
\begin{align}
    B_n &= \nabla g_n^\text{tail}({\beta_\ell,\beta_\text{u}}) - \nabla g_n^\text{tail}({\overline{\beta}_\ell,\overline{\beta}_\text{u}}) \\ &\quad+\nabla g_n^\text{head}({\beta_\ell,\beta_\text{u}}) - \nabla g_n^\text{head}({\overline{\beta}_\ell,\overline{\beta}_\text{u}}).
\end{align}
$(c)$ arises from the inequality 
\begin{equation}
\small
\begin{split}
    \left\Vert \sum_{n=1}^{N} B_n\right\Vert\leq \sum_{n=1}^{N}\left\Vert\nabla g_n^\text{tail}({\beta_\ell,\beta_\text{u}}) - \nabla g_n^\text{tail}({\overline{\beta}_\ell,\overline{\beta}_\text{u}})\right\Vert \\+\left\Vert\nabla g_n^\text{head}({\beta_\ell,\beta_\text{u}}) - \nabla g_n^\text{head}({\overline{\beta}_\ell,\overline{\beta}_\text{u}})\right\Vert.
    \end{split}
\end{equation}
Since the Lipschitz constants in \eqref{constraint lipschitz} do not depend on specific data, we extend and generalize the preceding inequality into the constraint function:
\begin{equation}\label{constraintgradientcontinuous}
  \left\Vert\nabla v(\beta_{\ell}, \beta_{\text{u}}) - \nabla v(\overline{\beta}_{\ell},\overline{\beta}_{\text{u}}) \right\Vert \leq 2DM\gamma\cdot\left\Vert \bm\beta\right\Vert.
\end{equation}
In this case, the constraint function's gradient is demonstrated to be Lipschitz continuous with the Lipschitz constant $2DM\gamma$.
According to the lemma \ref{gradientcontinuousobject}, it is straightforward that
\begin{align}\label{energy1}
    &\left\Vert\frac{P_{\text{tr}}D}{{R_\text{tr}}}\sum_{n=1}^{N}\left(\frac{\mathbb{I}^\text{tail}_{n}(\beta_\ell,\beta_{\text{u}})}{\partial\beta_\ell\partial \beta_\text{u}}- \frac{\mathbb{I}^\text{tail}_{n}(\overline{\beta}_\ell,\overline{\beta}_{\text{u}})}{\partial\overline{\beta}_\ell\partial \overline{\beta}_\text{u}}\right)\right\Vert \\&\leq  \frac{k^2N(N+1)(N+4\sqrt{3}-1)P_{\text{tr}}D}{24R_\text{tr}}\cdot\left\Vert\bm\beta\right\Vert,
\end{align}
Let $o(\beta_\ell,\beta_\text{u}) = \sum_{n=1}^{N}[ \mathbb{I}^\text{tail}_{n}(\beta_\ell,\beta_{\text{u}})+ \mathbb{I}^\text{head}_{n}(\beta_\ell,\beta_{\text{u}})]E_\text{loc}(n)$, it follows (\ref{constraint lipschitz}) that
\begin{equation}\label{energy2}
\small
\begin{split}
     &\left\Vert \frac{\partial o(\beta_\ell,\beta_\text{u})}{\partial\beta_\ell\partial\beta_\text{u}} - \frac{\partial o(\overline{\beta}_\ell,\overline{\beta}_\text{u})}{\partial\overline{\beta}_\ell\partial\overline{\beta}_\text{u}}\right\Vert \\&\leq \sum_{n=1}^{N}2k^2(\frac{n^2-n}{8}+\frac{\sqrt{3}n}{3})E_\text{loc}(n)\cdot \left\Vert \bm\beta\right\Vert \\ &\leq \frac{k^2N(N+1)(N+4\sqrt{3}-1)}{12}E_\text{loc}(N)\cdot\left\Vert\bm\beta\right\Vert.
\end{split}
\end{equation}
Since the $f_\text{energy}$ contains the term in both \eqref{energy1} and \eqref{energy2} and the Lipschitz constant can be generalized to all $M$ events, $f_\text{energy}$'s gradient follows
\begin{equation}
\small
\begin{split}
     &\left\Vert\nabla f_\text{energy}(\beta_{\ell},\beta_{\text{u}}) - \nabla f_\text{energy}(\overline{\beta}_{\ell},\overline{\beta}_{\text{u}}) \right\Vert \\&\leq \frac{Mk^2N(N+1)(N+4\sqrt{3}-1)}{12}\left(E_\text{loc}(N)+\frac{P_{\text{tr}}D}{2R_\text{tr}}\right)\left\Vert \bm\beta\right\Vert.
     \end{split}
\end{equation}
According to the definition of weakly convexity and smoothness, $f_\text{energy}(\beta_\ell,\beta_{\text{u}})$ is proved to have its gradient Lipschitz continuous with the Lipschitz constant $2M\gamma\left(E_\text{loc}(N)+\frac{P_{\text{tr}}D}{2R_\text{tr}}\right)$.
\noindent Proof of Lemma \ref{energycontinuity} is completed.

\vspace{-3mm}
\section{Proof of Lemma \ref{convexandsmooth}}\label{proposition2}
Given $\bm\beta = \begin{pmatrix}
\beta_\ell-\overline{\beta}_\ell \\
   \beta_\text{u}-\overline{\beta}_\text{u}
\end{pmatrix}$, since both objective and constraint functions are proved to follow continuous Lipschitz gradient in previous two Lemmas. According to Lemma 1.2.3 from Nesterov's work \cite{convex}, it is straightforward to prove:
\begin{equation}\label{objectconvex}
\small
 \left\Vert f_{\text{acc}}(\beta_\ell,\beta_\text{u}) - f_{\text{acc}}(\overline{\beta}_\ell,\overline{\beta}_\text{u})-\left<\nabla f_{\text{miss}}(\overline{\beta}_\ell,\overline{\beta}_\text{u}),\bm\beta\right>\right\Vert\leq\frac{\gamma}{2}\left\Vert\bm\beta\right\Vert^2,
\end{equation}
and 
\begin{equation}\label{constraintconvex1}
\small
\left\Vert v(\beta_{\ell},\beta_{\text{u}}) - v(\overline{\beta}_{\ell},\overline{\beta}_{\text{u}})-\left<\nabla v(\overline{\beta}_{\ell},\overline{\beta}_{\text{u}}),\bm\beta\right>\right\Vert\leq DM\gamma\left\Vert\bm\beta\right\Vert^2, 
\end{equation}
\begin{equation}\label{constraintconvex2}
\small
\begin{split}
    &\left\Vert f_{\text{energy}}(\beta_\ell,\beta_\text{u}) - f_{\text{energy}}(\overline{\beta}_\ell,\overline{\beta}_\text{u})-\left<\nabla f_{\text{energy}}(\overline{\beta}_\ell,\overline{\beta}_\text{u}),\bm\beta\right>\right\Vert \\&\leq M\gamma\left(E_\text{loc}(N)+\frac{P_{\text{tr}}D}{2R_\text{tr}}\right)\left\Vert\bm\beta\right\Vert^2.
\end{split}
\end{equation}
Hence, \eqref{objectconvex} indicates that $f_{\text{acc}(\beta_\ell, \beta_\text{u})}$ is both $\gamma$-smooth and $\gamma$-weakly convex. \eqref{constraintconvex1} and \eqref{constraintconvex2} implies that $v(\beta_{\ell},\beta_{\text{u}})$ is both $2DM\gamma$-smooth and $2DM\gamma$-weakly convex, $f_\text{energy}(\beta_\ell, \beta_\text{u})$ exhibits $2M\gamma\left(E_\text{loc}(N) + \frac{P_{\text{tr}}D}{2R_\text{tr}}\right)$ weak convexity and smoothness.
\noindent Proof of Lemma \ref{convexandsmooth} is completed.

\vspace{-2mm}
\section{Proof of Proposition \ref{energyipppconvexsmooth}}\label{Proposition3}
Given $\bm\beta = \begin{pmatrix}
\beta_\ell-\overline{\beta}_\ell \\
   \beta_\text{u}-\overline{\beta}_\text{u}
\end{pmatrix}$. Since $f_\text{acc}(\beta_{\ell},\beta_{\text{u}})$ has proved its weakly convexity,  $P_\text{off}$ follows that:
\begin{equation}\label{offloadinginequality}
    P_\text{off}(\beta_{\ell},\beta_{\text{u}})\ge P_\text{off}(\overline{\beta}_{\ell},\overline{\beta}_{\text{u}}) + \left\langle\nabla P_\text{off}(\overline{\beta}_{\ell},\overline{\beta}_{\text{u}}), \bm\beta\right\rangle - \gamma \left\Vert\bm\beta\right\Vert^2.
\end{equation}
Based on \eqref{offloadinginequality} and the fact that $\frac{\kappa}{2}(\max \{0, P_\text{off}(\beta_{\ell},\beta_{\text{u}})\})^2$ is still weakly convex, it is straightforward to derive:
\begin{equation}
\small
\begin{split}
   &\frac{\kappa}{2}(P_\text{off}(\beta_{\ell},\beta_{\text{u}}))^2 \\&\ge \frac{\kappa}{2}P_\text{off}(\overline{\beta}_{\ell},\overline{\beta}_{\text{u}}))^2+\kappa P_\text{off}(\overline{\beta}_{\ell},\overline{\beta}_{\text{u}})\cdot\left(P_\text{off}(\beta_{\ell},\beta_{\text{u}})-P_\text{off}(\overline{\beta}_{\ell},\overline{\beta}_{\text{u}})\right) \\ &\ge \frac{\kappa}{2}P_\text{off}(\overline{\beta}_{\ell},\overline{\beta}_{\text{u}}))^2+ \kappa P_\text{off}(\overline{\beta}_{\ell},\overline{\beta}_{\text{u}})\cdot \left(\left\langle\nabla P_\text{off}(\overline{\beta}_{\ell},\overline{\beta}_{\text{u}}), \bm\beta\right\rangle - \gamma \left\Vert\bm\beta\right\Vert^2\right).
\end{split}
\end{equation}
Let $x = ({\beta}_{\ell},{\beta}_{\text{u}})$ and $y = (\overline{\beta}_{\ell},\overline{\beta}_{\text{u}})$ for simplicity. Therefore, it is straightforward to show
\begin{equation}\label{A's expression1}
    \frac{\kappa}{2}(P_\text{off}(x))^2-\frac{\kappa}{2}P_\text{off}(y))^2\ge  \kappa P_\text{off}(y)\cdot \left\langle\nabla P_\text{off}(y), \bm\beta\right\rangle - \kappa A\gamma \left\Vert\bm\beta\right\Vert^2,
\end{equation}
By applying the triangle inequality and using the smoothness of $P_\text{off}(\beta_{\ell},\beta_{\text{u}})$, the gradient of $\frac{\kappa}{2}\left(P_\text{off}\left(\beta_{\ell},\beta_{\text{u}}\right)\right)^2$ is Lipschitz continuous as
\begin{equation}
\begin{split}
&\nabla \| \frac{\kappa}{2}(P_\text{off}(x))^2 - \frac{\kappa}{2}(P_\text{off}(y))^2\|
\\&= \kappa\| P_\text{off}(x)\nabla P_\text{off}(x) - P_\text{off}(y)\nabla P_\text{off}(y)\| \\ &= \kappa \left(|P_{\text{off}}(x) - P_{\text{off}}(y)| \|\nabla P_{\text{off}}(x)\|\right) \\&+ \kappa\left(|P_{\text{off}}(y)| \|\nabla P_{\text{off}}(x) - \nabla P_{\text{off}}(y)\|\right)
\end{split}
\end{equation}
Since $\|\nabla P_{\text{off}}(x) - \nabla P_{\text{off}}(y)\|\leq 2\gamma \left\Vert\bm\beta\right\Vert^2$, it follows that
\begin{equation}\label{A's expression2}
       \|\nabla  \frac{\kappa}{2}(P_\text{off}(x)^2 - \nabla \frac{\kappa}{2}P_\text{off}(y)^2\| \leq \kappa A (A + 2\gamma) \|x - y\|,
\end{equation}
For both \eqref{A's expression1} and \eqref{A's expression2}, $A$ corresponds to the constant that
\begin{equation}\label{constantA}
\small
\begin{split}
    A &\ge \max\left\{\left|P_\text{off}(\beta_\ell,\beta_\text{u})\right|,\left\Vert\nabla P_\text{off}(\beta_\ell,\beta_\text{u})\right\Vert\right\} \\&= \max\left\{\theta,\frac{DM(N-1)}{2\sqrt{2}}\right\} .
\end{split}
\end{equation}
In this case, $\frac{\kappa}{2}(P_\text{off}(x))^2$ is proved as a $2\kappa \gamma A$ weakly convex and $\kappa A\left(A + 2\gamma\right)$ smooth function. 
Since $v(\beta_{\ell},\beta_{\text{u}}) = DMP_\text{off}(\beta_{\ell},\beta_{\text{u}})$, it is straightforward that $\frac{\kappa}{2}(\max \{0, v(\beta_{\ell},\beta_{\text{u}})\})^2$ is a $2\kappa DM\gamma A$ weakly convex and $\kappa DMA\left(A + 2\gamma\right)$ smooth function. Given the proved fact that
$f_\text{energy}(\beta_\ell, \beta_\text{u})$ exhibits $2M\gamma\left(E_\text{loc}(N) + \frac{P_{\text{tr}}D}{2R_\text{tr}}\right)$ weak convexity and smoothness from \eqref{constraintconvex2}, similar mathematical derivation can done according to \eqref{A's expression1} and \eqref{A's expression2}. In this case,  $\frac{\rho}{2}\left(\max \{ 0,f_\text{energy}(\beta_\ell,\beta_\text{u})\}\right)^2$ exhibits $\rho B\left(B + 2M\gamma \left(E_\text{loc}(N) + \frac{P_{\text{tr}}D}{2R_\text{tr}}\right)\right)$ smoothness and $2M\gamma \rho B\left(E_\text{loc}(N) + \frac{P_{\text{tr}}D}{2R_\text{tr}}\right)$ weak convexity. Specifically, the Lipschitz constant is $2\rho B$ when $B$ is chosen to satisfy:
\begin{equation}
\begin{split}
    B &\ge \max\left\{\left|f_\text{energy}(\beta_\ell,\beta_\text{u})\right|,\left\Vert\nabla f_\text{energy}(\beta_\ell,\beta_\text{u})\right\Vert\right\} \\&= \max\left\{\xi, \frac{(N^2+1)E_\text{loc}(N)}{2\sqrt{2}}+\frac{(N+2)(N-1)P_\text{tr}D}{4\sqrt{2}R_\text{tr}}\right\}.
\end{split}
\end{equation}
\noindent Proof of Proposition \ref{energyipppconvexsmooth} is completed.

\bibliographystyle{IEEEtran}
\bibliography{Ref}
\end{document}